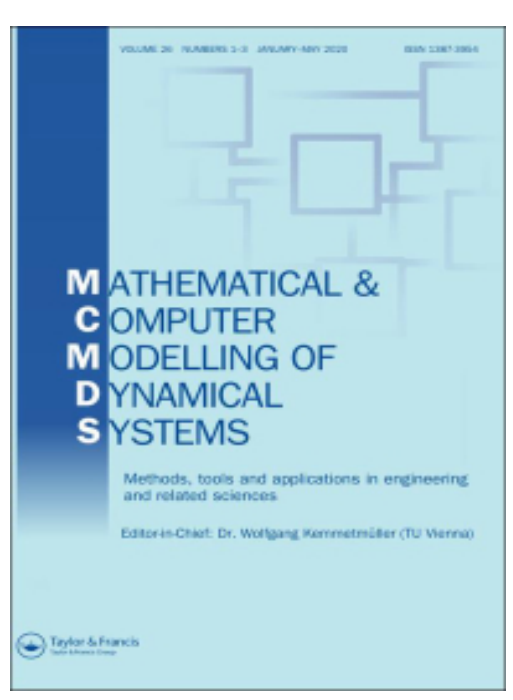

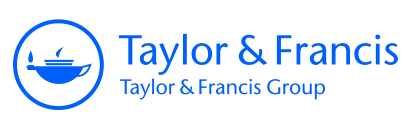

# RoboWalk: augmented human-robot mathematical modelling for design optimization

S. Ali A. Moosavian, Mahdi Nabipour, Farshid Absalan & Vahid Akbari

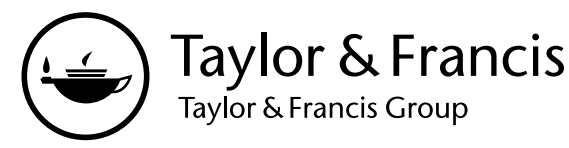

ARTICLE

OPEN ACCESS 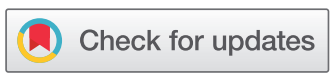

# RoboWalk: augmented human-robot mathematical modelling for design optimization

S. Ali A. Moosavian, Mahdi Nabipour, Farshid Absalan and Vahid Akbari

Centre of Excellence in Robotics and Control, Advanced Robotics & Automated Systems (ARAS) Lab, Department of Mechanical Engineering, K. N. Toosi University of Technology, Tehran, Iran

**ABSTRACT**

Utilizing exoskeleton devices to help elderly or empower workers is a growing field of research in robotics. The structure of an exoskeleton can vary depending on user's physical dimensions, joints or muscles targeted for assistance, and maximum achievable actuator torque. In this research, a Human-Model-In-the-Loop (HMIL) constrained optimization technique is proposed to design the RoboWalk lower-limb exoskeleton. RoboWalk is an under-actuated non-anthropomorphic assistive robot, that besides applying the desired assistive force, exerts an undesirable disturbing force leading to the user's fall. The HMIL method uses the augmented human-robot 2D model to take RoboWalk and human body's joint torques into account during optimization. The superiority of HMIL method is proven by comparing the results with other strategies in the literature. Obtained results reveal elimination of the disturbing forces, 2 N.m. reduction in average human knee-joint torque, and significant decrease in the actuator required torque.

## Introduction

Humanoids are inherently unstable robots that should be dynamically stabilized [1], especially when performing tasks on different surface conditions [2,3]. Human-robot rehabilitation technology is one of the spinoffs of humanoid technology where, depending on the user's health, a part of the tasks is taken over by the human itself [4]. Orthosis and exoskeletons are wearable robots that are widely used in medical industries for curing patients suffering from limb pathology and for performance augmentation of able-bodied users. The medical exoskeletons are categorized into two groups of treadmill-based immobile [5–7] and mobile devices [8–10]. The BLEEX exoskeleton [11] of University of California, Berkeley is the most well-known exoskeleton of strength augmentation type. Robo-Knee [12] and MIT augmentation exoskeleton [13] are other types of these devices.

**CONTACT** Mahdi Nabipour m.nabipour@email.kntu.ac.ir Centre of Excellence in Robotics and Control, Advanced Robotics & Automated Systems (ARAS) Lab, Department of Mechanical Engineering, K. N. Toosi University of Technology, Tehran, Iran

### *Assistive wearable robots*

Another important class of wearable robots are assistive exoskeletons. The intention of using such groups of exoskeletons is assisting the weak elderly to become needless of walker, sticks or help of others in doing their daily activities. In addition, it can help healthy users to do their tasks by spending less energy in longer periods of time. This kind of exoskeletons are divided into two categories: passive and active. Passive devices use compliant elements instead of using any form of electrical power to operate and assist the user. As for some examples of passive assistive exoskeletons, one can mention the XPED2 [14], Passive Ankle Exoskeleton [15], MoonWalker [16].

One of the most successful active assistive robots is the Honda bodyweight support assist device [17]. The elders, tourists, factory workers and, generally, people who should be on foot for hours are the main target society of this device. This device has an attractive design such that wearing it is as easy as putting on the shoes and placing the seat under the groin region. The objective of the device control strategy is to offer maximum reduction in felt weight in single support phase. Although the results showed a 10% reduction in the Floor Reaction Force (FRF) during the mid-stance phase, unfortunately, an unwanted horizontal force in sagittal plane [18] was produced. This disturbing force causes interruptions and may be a threat to user stability or may increase joint torques during the walking gait cycle.

The design and function of Honda's bodyweight system has captured a lot of attention. The passive Bodyweight support exoskeleton with compliant knee [19] and assistive exoskeleton device fabricated in Ottawa University [20] are examples of assistive devices inspired by Honda bodyweight support assist device. The former claims to reduce the knee joint torque up to 27% in some points of mid-stance phase but although it uses a compliant knee, its design is not optimized, and it has a bad influence on user gait. In the latter case, the kinematic results revealed that the device interrupts user gait and does not assist the user in most of the gait cycle. In both cases, no optimizations were performed and this could be a source of these bad influences on user gait.

### *Design optimization*

In the exoskeleton industry, online optimization techniques are more frequently used for finding optimal set of control algorithms [21] or set of control parameters [22,23] by using the prediction of future kinematics or dynamics [24,25]. On the other hand, exoskeleton design should be optimized offline and before it is constructed, unless the device is modular, soft or changeable [26] which is not the case in this paper.

Offline dimensional optimization has lots of applications in multi-body systems [27, 28, 29] and robotic industries. Shim et al. have designed a robotic gripper for the front-end module assembly process and optimized a six-bar linkage by a multi-objective optimization approach [30]. Pan et al. designed a scissor mechanism for load-carrying augmentation [31]. In [32] a length optimization of the scissor sides to minimize the transmitting errors between the input and output motions in walking were performed. In another study, a passive upper limb exoskeleton was designed [33] and its elastic elements' impedance were optimized by genetic algorithm. Authors in [34] designed pleated pneumatic artificial muscles to obtain a lightweight design by maximizing the compactness of the actuator.

In the field of lower limb exoskeletons, Kawale [35] performed optimization to achieve an energy-efficient design for a wearable lower-body exoskeleton mechanism in shipbuilding industry. Ortiz [26] used a starting point dependent Nelder–Mead method for optimizing the structure and parameters of a soft modular lower limb exoskeleton to make it energetically efficient. Random sampling within the parameter space was performed to prevent finding the local minima. In Wang et al. an optimization framework to obtain efficient and lightweight combination for the Mindwalker series elastic joint was proposed [36]. None of the above-mentioned articles used human model joint torques as a criterion for their optimization. This problem is going to be addressed in this paper, along with other key elements included in the optimization algorithm.

This paper introduces new results along with the author's previous works [37, 38, 39], wherein [37] a conceptual design of the robot was added to a skeletal Opensim model and some preliminary simulations and tests were performed. The works presented in [38,39] dealt with the modelling and performance analysis of the first and second generation of RoboWalk exoskeleton. In this paper, a 2D human skeletal model and the third generation of RoboWalk exoskeleton are modelled in sagittal plane. To this end, an initiative accurate kinematic model for RoboWalk is presented that requires less computational time than other methods presented in [38,39], which makes it suitable for optimization. Also, for the first time, the Human-robot models are augmented to be used in the proposed HMIL strategy. The main goals of this human-robot model augmentation are to minimize the disturbance force, RoboWalk actuator torque and the user knee-joint torque.

The paper is divided into seven sections. In section 2, a brief introduction to RoboWalk assistive exoskeleton structure and its functionality are presented. The multibody augmented human-exoskeleton model is obtained by Newton and Recursive Newton–Euler Algorithm (RNEA) in section 3. In section 4, three constrained dimensional optimization techniques, including HMIL strategy, are presented. Section 5 presents the verification of the obtained dynamics and discusses the results obtained by the design optimization approaches by considering the human joint torques as a good criterion for analysing user comfort. Finally, the conclusions of this study are summarized in the last section.

## Introducing RoboWalk Assistive Device

As mentioned before, the structure of RoboWalk is inspired of Honda's bodyweight support assist device, which is explained in detail in [17,38] and demonstrated in Figure 1.

The user wears the shoes of the device and, after turning the device on, raises the seat up to the groin area. The front and rear bearings, the guide rails attached to the seat and the rollers inside it provide a passive 3-DoF rotational motion for the hip. The knees, which are actuated by the motor placed under the seat, have 1-DoF in sagittal plane and the ankle spherical joint permits a 3-DoF rotation. The torque generated by the actuator is tranfered to the knee joint by a 4-link transmission mechanism. There are some points worth noting about RoboWalk's motion:

(1) The legs of RoboWalk (extraction and contraction mechanisms) remain between the legs of the user during the entire gait. This is a beneficial design allowing the device to have a low moment of inertia when rotating about the user longitudinal axis.
(2) This design restricts the robot to move in the sagittal plane whilst the user is walking, therefore all the analysis is done in the sagittal plane.
(3) A minimal actuator torque (0–8 N.m.) is always applied to hold the seat attached to the buttocks. Therefore, during simulations the robot seat is assumed to be stuck to user pelvis.
(4) The robot's ankle joint is fixed to user shoes. Hence, after the robot is set in place, RoboWalk kinematics is completely specified.

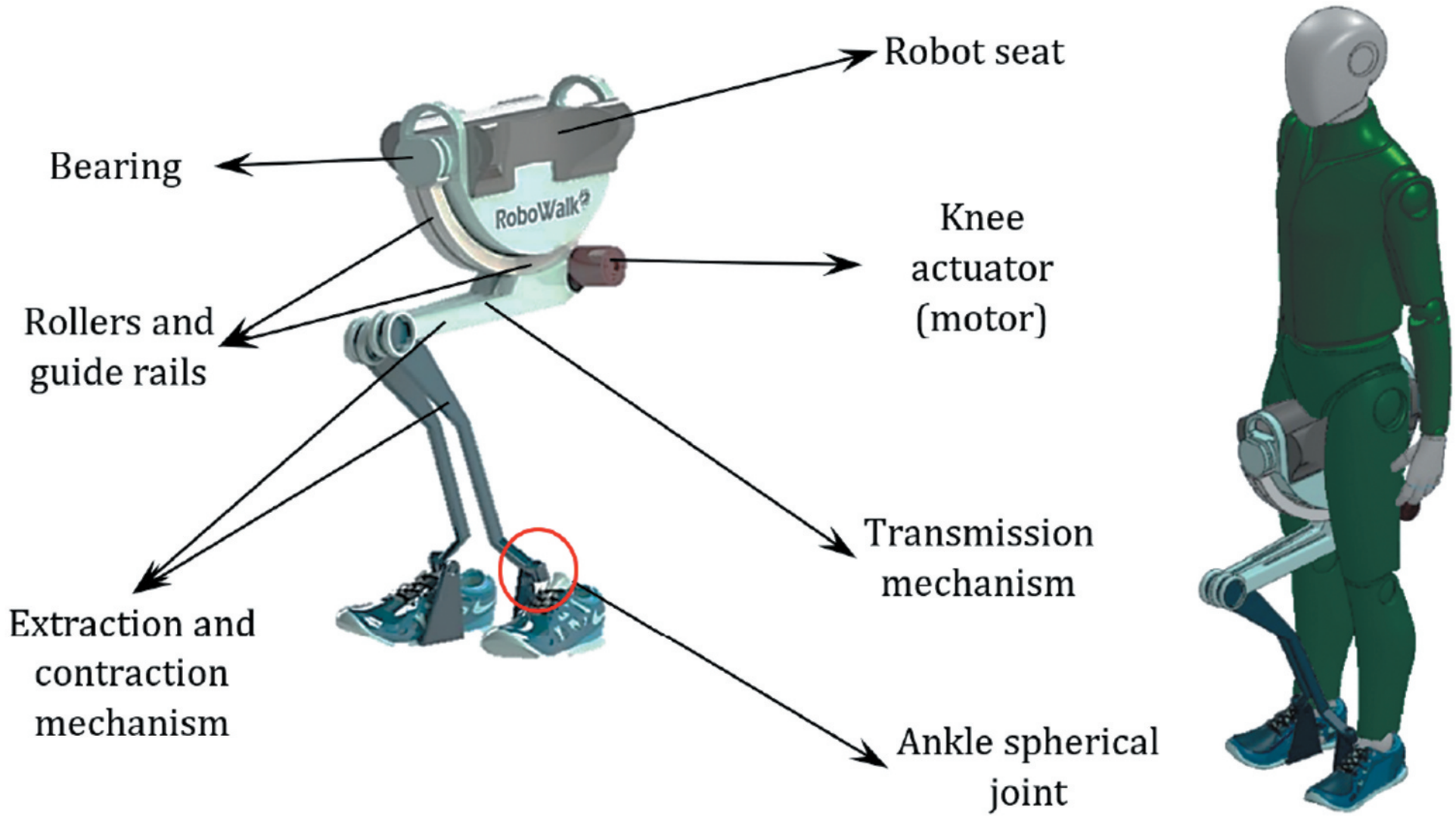

**Figure 1.** RoboWalk structure.

A closer look to the device structure (Figure 2) could be more illuminating.

Similar to Honda's assistive device, the desired upward assist force is set to be a constant value during the optimal design simulation. This assisting force is applied to the seat by the stance leg. The torque $T_m$ is approximately the same as the torque applied to the knee joint during the entire stance phase. Meanwhile, the knee actuator of the leg in the swing phase is inactive to avoid disturbing the swing leg. Therefore, in addition to applying the assisting force, the stance side of the device bares the weight of the robot's leg in the swing phase. The undesirable disturbing force exerted to the human arises when the stance leg tends to apply the assisting upward force. Due to the under-actuated nature of the device, a horizontal force is applied to the user as a side effect of the upward force.

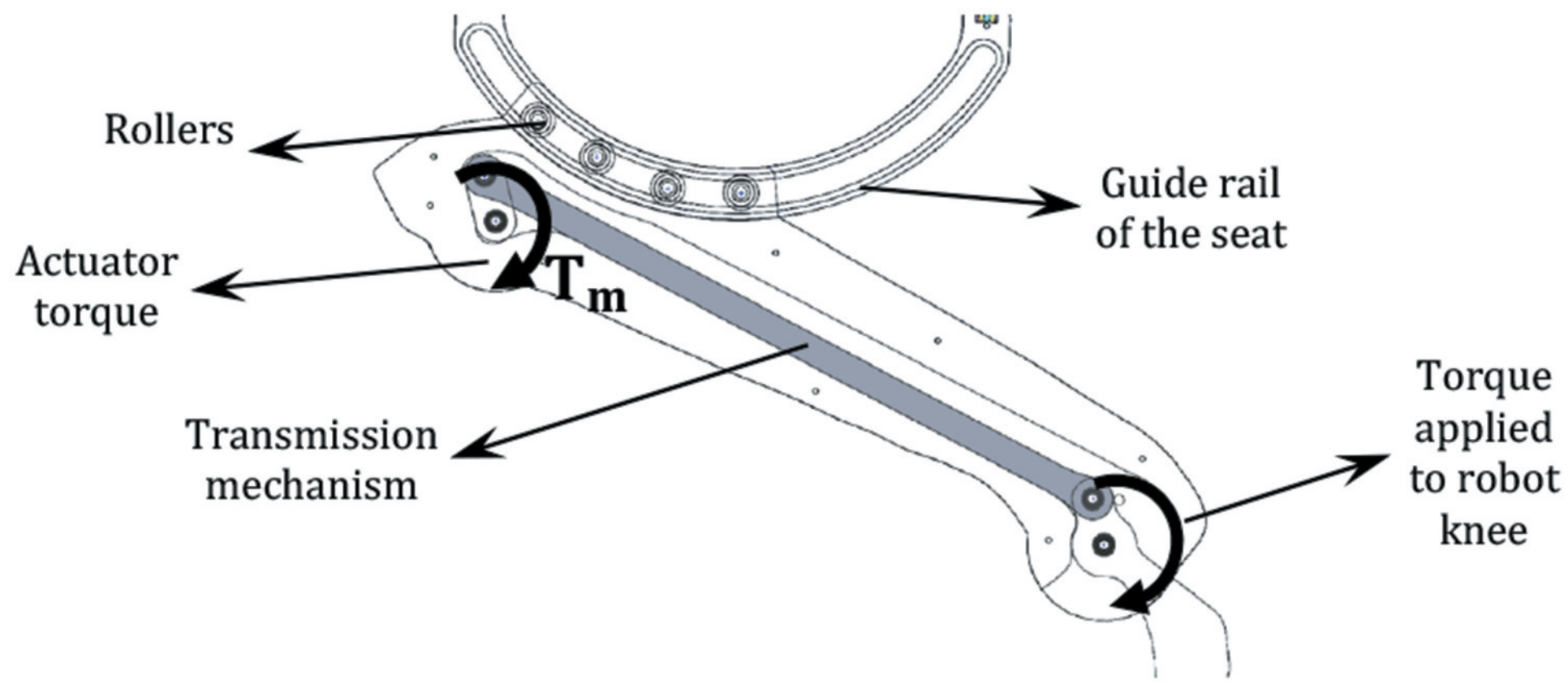

**Figure 2.** Upper link and seat of RoboWalk.

## Integrated System Mathematical Modelling

In this section, the kinematics and dynamics of the human skeletal model and RoboWalk are discussed. The procedure of deriving the equations of motion (EoM) of the system is depicted in Figure 3 briefly.

$X_p$ and $Z_p$ are the pelvis cartesian coordinates in sagittal plane and $q$ is the set of human trunk, pelvis, thigh, shank and foot angles. The first task in obtaining user's forward kinematics is to find the foot contact point. This could be done by finding the maximum height between a reference point (e.g., pelvis) and toe/heel of the user and assigning the maximum amount as the height of the reference point [38]. Since the connection points of RoboWalk are invariant with respect to the human body, by knowing the forward kinematics of the user, the inverse kinematics of RoboWalk can be derived. It is shown in the subsequent subsections that by knowing the robot's inverse dynamics, the robot's interaction forces with user are achieved.

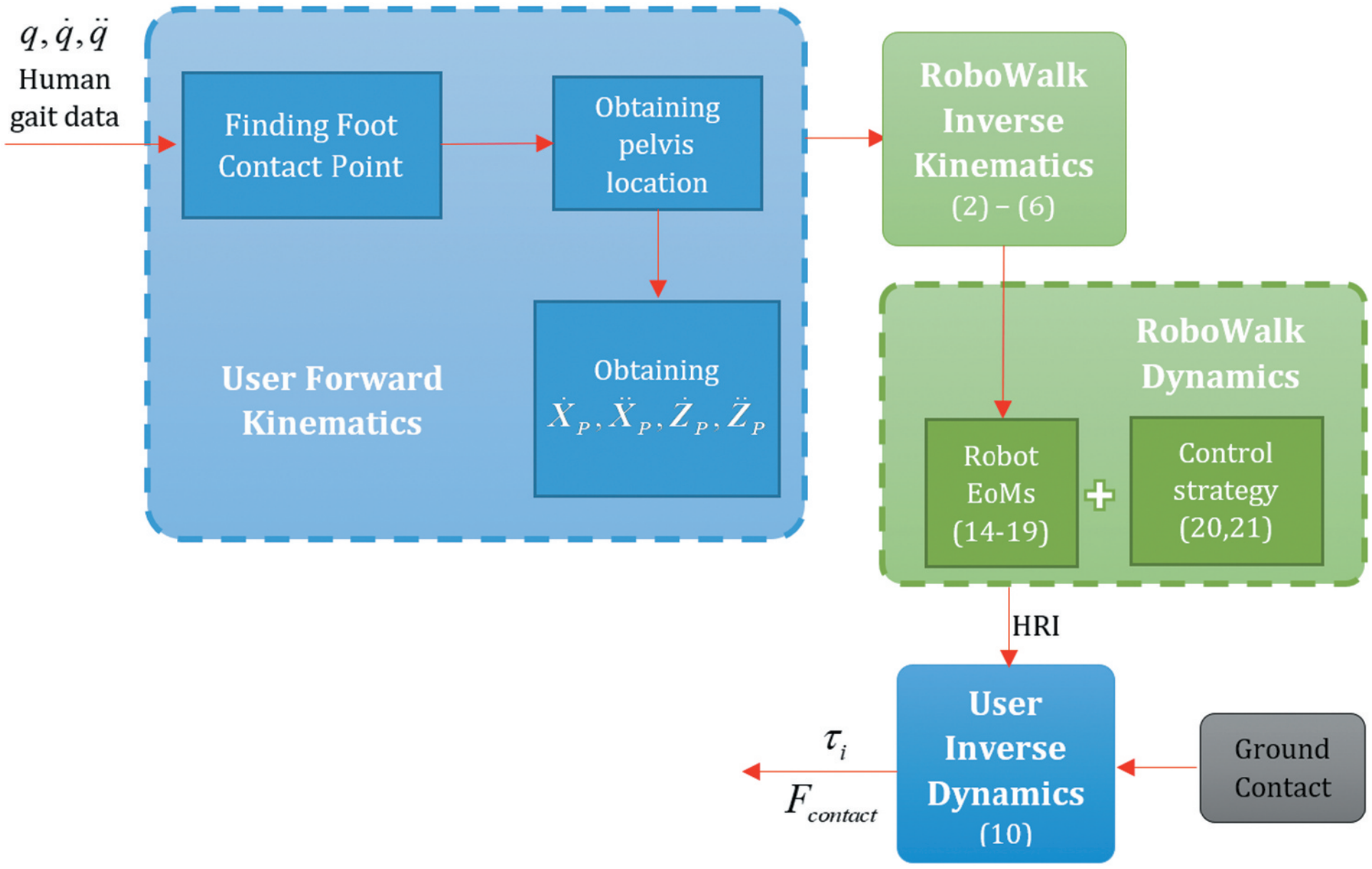

**Figure 3.** Procedure for deriving the EoMs of the system.

RoboWalk is connected to the user in three points: the seat and the two robot ankles that are connected to the user shoes. The interaction forces between a robot and a human are known as human-robot interactions (HRI). These forces along with the FRFs are assumed as external forces to the user inverse dynamics model. The details of this procedure and the governing equations are discussed in subsequent subsections.

## *Kinematics Modelling*

### *Human kinematic model*

Since RoboWalk is limited to be between the user's leg during the gait, the robot's movement is restricted to the sagittal plane and almost all of the human-robot interaction forces are limited to this plane. Therefore, although RoboWalk is free to move passively with the user in other planes (e.g. when the user performs hip abduction/adduction), since we are investigating human walking and RoboWalk's influence on that, all the modellings and simulations are done in sagittal plane. In addition, all the analysis is conducted in single support phase (SSP). Accordingly, the human model is assumed to be an 8 linked multi-body system which includes pelvis, trunk, right and left thigh, shank and foot. The upper extremity is replaced by the trunk for ease in analysis. The pelvis is assumed to be the floating-base. Thus, the pose and orientation of the coordinate system attached to the centre of pelvis is calculated with respect to the inertial coordinate system to allow movement analysis of the human user. All other joints are assumed as ideal rotary joints and the inertial coordinate system is located on the ground and under the user before starting to walk. All coordinate axes of this model are illustrated in Figure 4.

In this figure, the arrows are the x-direction of coordinate frames and the crosses are the z-direction. The vector of generalized coordinates that define the whole-body motion is specified as:

$$q = \begin{bmatrix} X_p & Z_p & q_p & q_t & q_h & q_k & q_a \end{bmatrix}^T \tag{1}$$

where $X_p$, $Z_p$ and$q_p$ are the pose and orientation of pelvis reference coordinate with respect to the inertial coordinate system, expressed in inertial reference. $q_t$, $q_h$, $q_k$ and $q_f$ represent the rotational motion of trunk, hip, knee and ankle joints for left and right legs, respectively. Since our analysis is restricted to sagittal plane, these joints represent 1-DoF revolute joints. By this definition, the human model is a 10-DoF kinematic tree structure in sagittal plane. The governing equations of human kinematics model are completely discussed in [38].

### *RoboWalk Kinematic Model*

As stated before, the robot interacts with the user in three points (both ankles and the seat). The directions of the forces applied to the seat are determined by kinematic calculations. Consider the human and robot as shown in Figure 5.

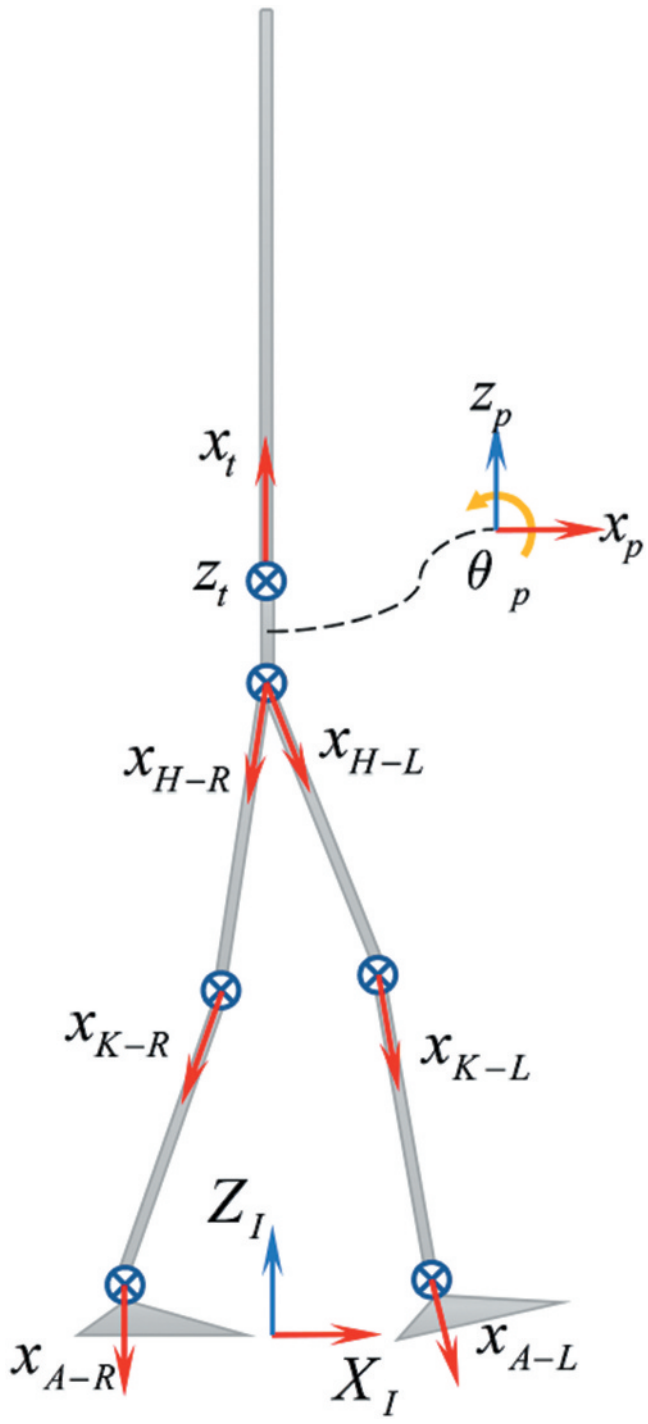

**Figure 4.** Coordinate systems used in the kinematics of user.

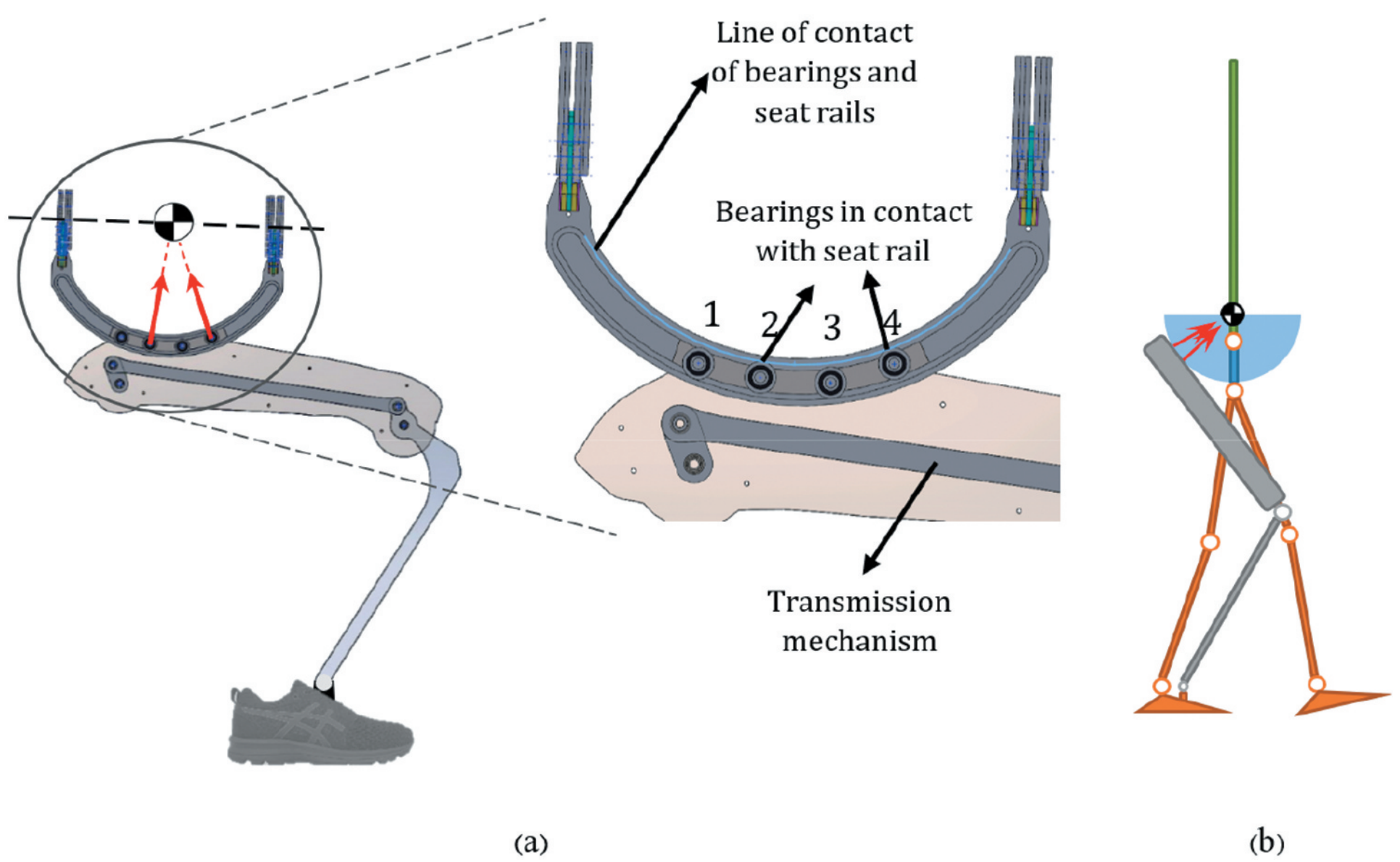

**Figure 5.** A) Force transmission to the user b) user and robot schematic.

As shown in Figure 5-a, wherever the roller bearings are in contact with the seat rails, a normal force is exerted to the rail (tangential friction force is neglected). These forces are transmitted to the user through the seat, in the horizontal and vertical directions, as the assisting upward force and friction horizontal force. The rollers are attached to the upper leg of the mechanism and are free to move within the guide rails. Rollers number 2 and 4 push the seat up and numbers 1 and 3 pull it down whenever needed. The forces exerted to the seat are directed towards the centre of the guide rail arc (which is assumed to be almost coincident with the user's centre of gravity (CoG)). Figure 5-b is a schematic of the user and one leg of the assistive robot and the direction of the assist force.

Since RoboWalk has a special structure, obtaining its kinematics is a bit tricky. Figure 6 demonstrates the important angles for introducing the kinematics of the system.

In this figure $(x_0, z_0)$is the centre of guide rail arcs (i.e. approximate CoG of user) and $(x_e, z_e)$ is the coordinate of robot ankle. The bearings attached to the upper limb are constrained to move inside the guide rail. Thus, the upper link is forced to rotate about the arc centre, which makes it possible to handle this complex constrained geometry and simplifies the leg's motion analysis. Accordingly, the entire leg kinematics model is simplified to a two-link manipulator with the lengths of $L$ and $r$ as depicted in Figure 6. Hence, the inverse kinematics of the robot in any configuration is as follows.

$$\theta_a = \tan^{-1}\left(\frac{x_e - x_0}{z_e - z_0}\right) \tag{2}$$

$$\theta_b = \cos^{-1}\left(\frac{L^2 + P^2 - r^2}{2LP}\right) \tag{3}$$

Since$\theta_1 = \theta_a + \theta_b$, the first link's angle with respect to vertical is known. By simple maths operations $\theta_2$ is obtained as follows.

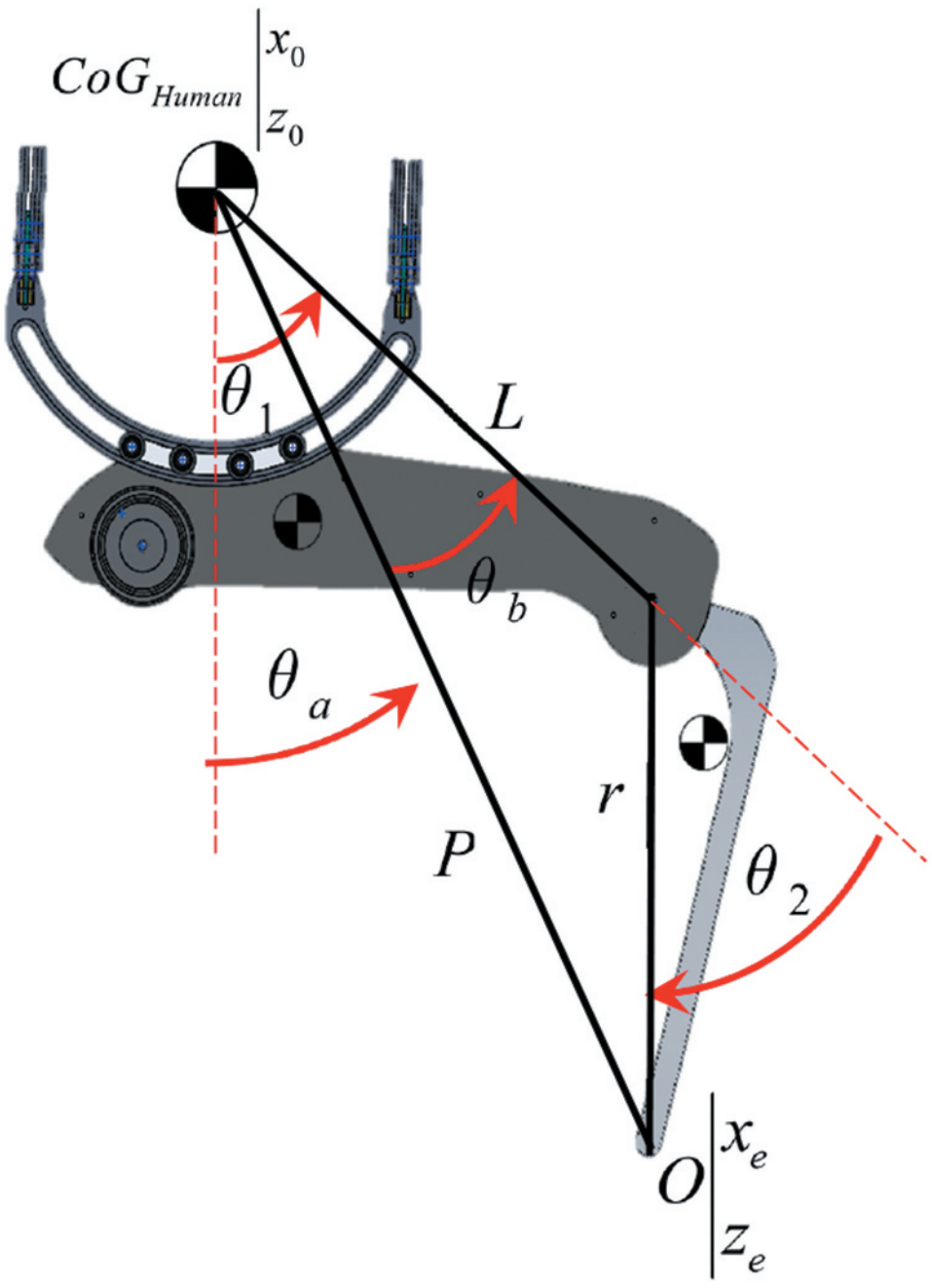

Figure 6. Geometrical dimensions required for the kinematics of RoboWalk.

$$\theta_2 = \cos^{-1}\left(\frac{P^2 - L^2 - r^2}{2Lr}\right) \tag{4}$$

By knowing the velocity and acceleration of $(x_0, z_0)$ and $(x_e, z_e)$ and introducing $\Delta x = x_e - x_0$, $\Delta z = z_e - z_0$and $\Theta = \begin{bmatrix} \theta_1 & \theta_2 \end{bmatrix}^T$, the links' angular velocities and accelerations are computed as follows.

$$\dot{\Theta} = J^{-1}\begin{bmatrix} \Delta\dot{x} \\ \Delta\dot{z} \end{bmatrix} \tag{5}$$

$$\ddot{\Theta} = J^{-1}\left(\begin{bmatrix} \Delta\ddot{x} \\ \Delta\ddot{z} \end{bmatrix} - \dot{J}\dot{\Theta}\right) \tag{6}$$

where $J$ represents the jacobian of the robot ankle with respect to human's CoG.

## Augmented Dynamics Modelling

After obtaining the kinematics relations of human model and RoboWalk, and possessing the gait data from Opensim [40] software, the inverse dynamics relations can be acquired. In order to obtain the human-RoboWalk integrated model, the HRI forces are assumed as external forces applied to the human model. The impact of the HRI forces on the human skeletal model is observed by using the transpose of jacobian of the contact points.

## Human-Body Dynamics model

As stated before, the FRF and HRI forces are the main external forces applied to the user. Hence, a general form of user dynamics is obtained and these external forces are then introduced to the equation. The general equation of a system, when no constraints and external forces are applied to, is expressed as:

$$M(q)\ddot{q} + C(q, \dot{q}) + G(q) = Q \tag{7}$$

in this equation, $M$ represents a $10 \times 10$ inertia matrix, $C$ and $Q$ are $10 \times 1$ vectors. $C$ represents Coriolis, centrifugal, gyration effects, $G$ represents the gravity effects and $Q$ is equivalent to $D\tau$ when the human is considered as a floating body with no interaction with the environment [41]. In this case, $D$ and $\tau$ are defined as:

$$D = \begin{bmatrix} 0_{3\times7} \\ I_{7\times7} \end{bmatrix}$$
$$\tau = \begin{bmatrix} \tau_T & \tau_{HR} & \tau_{KR} & \tau_{AR} & \tau_{HL} & \tau_{KL} & \tau_{AL} \end{bmatrix}^T \tag{8}$$

in which$\tau_T$is the trunk joint torque,$\tau_{HR}, \tau_{KR}, \tau_{AR}$ are joint torques of right hip, knee and ankle and $\tau_{HL}, \tau_{KL}, \tau_{AL}$are those of the left leg, respectively. Then, in order to include the external ground contact forces in the EoM, one could develop constrained dynamics model by simply replacing unknown forces/moments acting on the point of contact. The mapping of these forces to the space of generalized coordinates is carried out using the transpose of jacobian of the contact point. In this case, $Q = D\tau + {J_{FRF}}^T F_{contact}$ where $J_{FRF}$ is the jacobian of contact point and $F_{contact}$ is the vector of external forces/moments exerted to that point. Thus, the EoM of the human model can be specified as:

$$M(q)\ddot{q} + C(q, \dot{q}) + G(q) = D\tau + J_{FRF}{}^{T} F_{contact} \tag{9}$$

When RoboWalk is exploited, the HRI forces are added to EoM as external forces. These HRI forces consist of the force exerted to the foot of the user by robot's ankle joint and the assisting forces applied to user's pelvis towards user CoG. Hence, the EoM of the augmented human-robot in SSP turns into the following equation.

$$[\tau F_{contact}]_{10\times 1} = DJ_{FRF}{}^{T} - 110 \times 10\left[M(q)\ddot{q} + C(q, \dot{q}) + G(q) - J_A{}^{T} F_A - J_s{}^{T} F\right] \tag{10}$$

where$J_s$ and $F$ are the jacobian of seat force applying point and the known assistive (including seat friction) forces obtained from robot inverse dynamics. In addition, in this equation

$$J_A{}^{T} F_A = J_{A_L} F_{A_L} + J_{A_R} F_{A_R} \tag{11}$$

where, $J_{A_L}$and $J_{A_R}$ are the robot's left and right ankle jacobian and $F_{A_L}$ and $F_{A_R}$ are the known forces applied to RoboWalk ankle joint.

Since $DJ_{FRF}{}^{T}$ is a square matrix, joint torques and contact forces are uniquely determined in SSP. In the double support phase (DSP) the EoM is as follows.

$$\begin{aligned} &M(q)\ddot{q} + C(q, \dot{q}) + G(q) = \\ &D\tau + J^{T}_{FRF_L} F_{FRF_L} + J^{T}_{FRF_R} F_{FRF_R} \end{aligned} \tag{12}$$

$J^{T}_{FRF_L}$and $J^{T}_{FRF_R}$ represent the Jacobian of the left and right foot contact points, respectively. In this case, the number of unknowns exceeds the number of EoMs by three (because the simulation is conducted in sagittal plane) and the problem is solved by Minimum Norm method.

$$A = A.\left(A^{T} A\right)^{-1} \tag{13}$$

Since the most extreme phase in designing RoboWalk is the SSP, and the fact that DSP is a very short time in human gait, the DSP is excluded from our analysis.

The dynamics equation of a multi-body system (Equation (7)) can be obtained by two different approaches. In the first approach, internal forces are omitted from the EoM (e.g. Lagrange method). in the second approach, the internal forces in the EoM are explicitly included (e.g. Newton–Euler method). Since the human joint force is a criterion of assessing the suitability of RoboWalk, Newton–Euler method is used for attaining the EoM of the entire system. The obtained system dynamics model is then verified using RNEA [42].

### *RoboWalk Dynamics Model*

After obtaining the kinematics of RoboWalk, the device's dynamics equation is obtained by assuming that two of the bearings are in contact with the guide rail in every instant. Figure 7 shows the FBD of the schematic assistive robot.

$m_1$ and $m_2$ are the masses of robot links. $O_1$ and $O_2$ are the centre of rotation of the upper and lower link, respectively. $F_1$ and $F_2$ represent the forces exerted by RoboWalk (roller bearings) to the user through the seat. $T_m$, $A_x$ and $A_z$ are the moment/forces in the device's knee. $B_x$ and $B_z$ are the forces applied to the user shoe by RoboWalk ankle. Similar to the method for obtaining human model dynamics, three equations could be obtained for each link. Equations obtained for upper link are

$$F_1 \cos \alpha_{f_1} + F_2 \cos \alpha_{f_2} + A_x = m_1 \ddot{x}_{G1} \tag{14}$$

$$F_1 \sin \alpha_{f_1} + F_2 \sin \alpha_{f_2} + A_z - m_1 g = m_1 \ddot{z}_{G1} \tag{15}$$

$$m_1 g a + A_x L \cos \theta_1 + A_z L \sin \theta_1 + T_m = I_{G1} \ddot{\theta}_1 + m_1 (b \ddot{x}_{G1} - a \ddot{z}_{G1}) \tag{16}$$

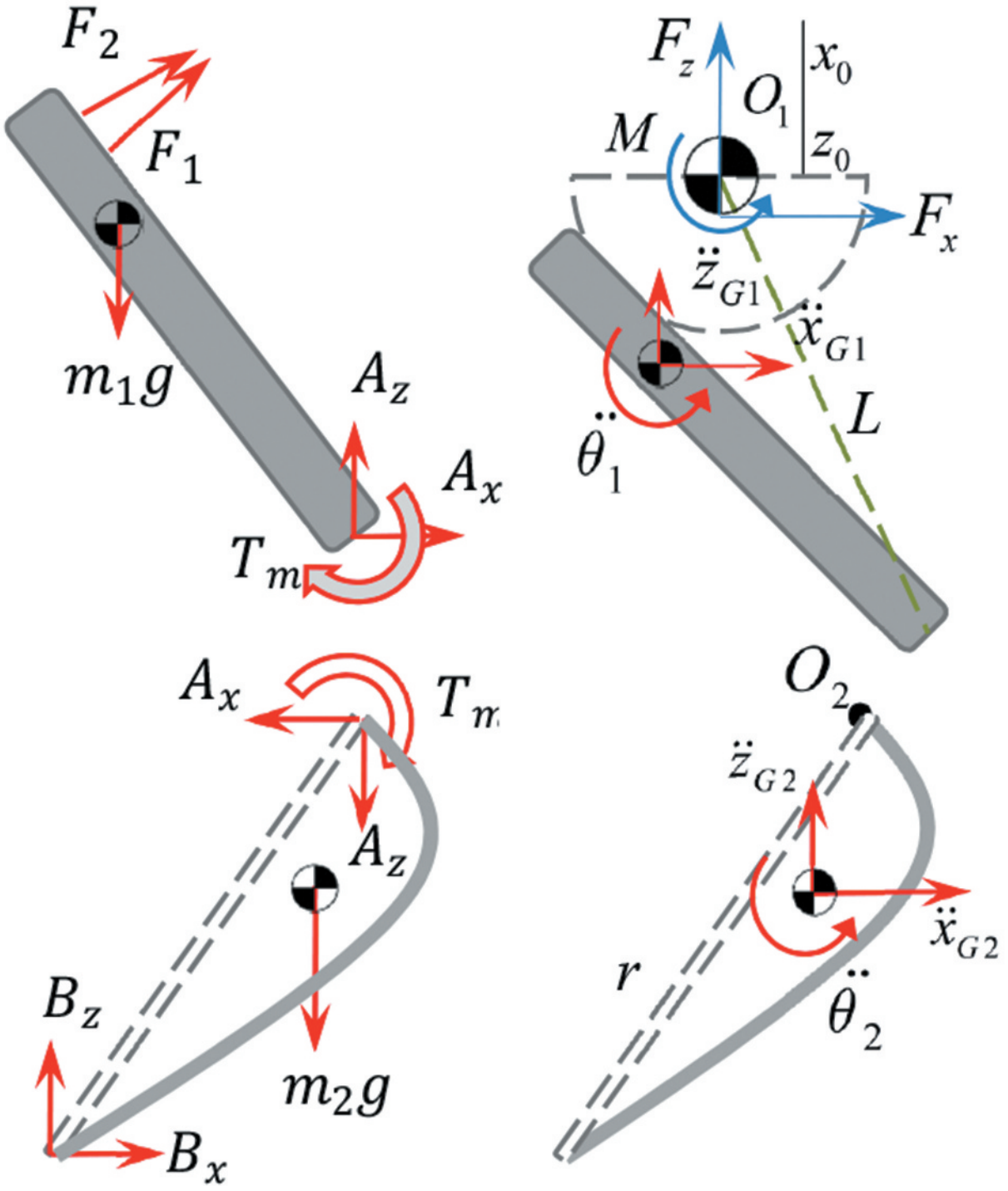

**Figure 7.** RoboWalk FBD and kinematic diagram in the sagittal plane.

Where $\alpha_{f_1}$ and $\alpha_{f2}$are the angle of applied forces with respect to horizontal. $a$ and $b$ are the horizontal and vertical distance between upper link CoG and human CoG ($O_1$) and $L$ is the distance between RoboWalk knee joint and human CoG ($O_1$). The same is done for the lower link as follow

$$B_x - A_x = m_2 \ddot{x}_{G2} \tag{17}$$

$$B_z - A_z - m_2 g = m_2 \ddot{z}_{G2} \tag{18}$$

$$\begin{aligned} &B_x r\cos(\theta_1+\theta_2)+B_z r\sin(\theta_1+\theta_2)\\ &-T_m-m_2 g r_{G2}\sin(\theta_1+\theta_2+\beta)=I_{G2}\left(\ddot{\theta}_1+\ddot{\theta}_2\right)\\ &+m_2\ddot{x}_{G2}r_{G2}\cos(\theta_1+\theta_2+\beta)+m_2\ddot{z}_{G2}r_{G2}\sin(\theta_1+\theta_2+\beta) \end{aligned} \tag{19}$$

where $r_{G2}$represent the distance between RoboWalk knee joint and lower link CoG. Hence, six equations are obtained for each leg of the robot. The vertical and friction force and the point where they are applied are found by writing the EoMs for the seat.

*Control Strategy*. As it is illustrated in Figure 8, there are seven unknown parameters ($\vec{F}_1, \vec{F}_2, A_x, A_z, B_x, B_z$and $T_m$) for each robot leg. Note that $\vec{F}_1$and $\vec{F}_2$directions are known by knowing the kinematics of the robot. Hence, without introducing another equation to these 6 equations, finding an accurate solution to this problem is impossible.

To achieve this goal, a control strategy should be defined for the assistive exoskeleton. In this paper, the main purpose of using the device is to support a portion of the user's bodyweight and induce the feeling of having less weight to the user. In order to do so, the objective of the robot stance leg is to compensate a portion of the human's bodyweight; the other leg's actuator is assumed to be turned off. Hence, the 7$^{th}$ equation for the stance leg is going to be:

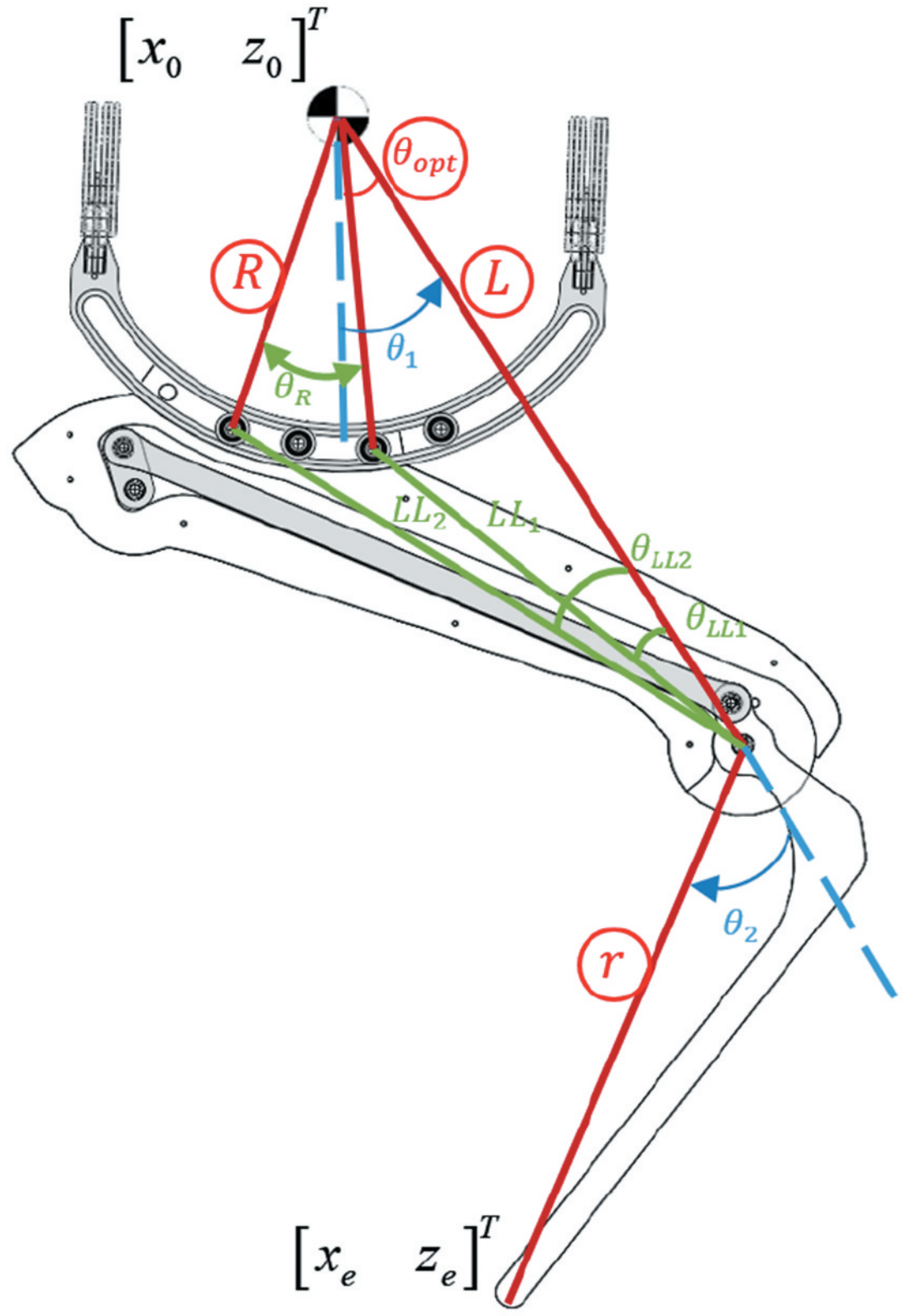

**Figure 8.** RoboWalk optimization parameters (circled parameters are directly obtained by optimization).

$$F_{z1} + F_{z2} = p\,.\,W_{human} \tag{20}$$

where $W_{human}$ is the total user weight and $p.W_{human}$ represents the portion of the bodyweight that the robot supports. For the leg in swing phase the 7th equation is:

$$T_m = 0 \tag{21}$$

By attaining these equations, all the unknown parameters are easily obtained in each sampling time.

### Recursive Newton–Euler Algorithm (RNEA)

RNEA is a numerical approach to model inverse dynamics. This algorithm is based on using 6 × 1 spatial coordinate [42] instead of the classical 3 × 1 representation [43] for each moment and force (or linear and angular velocity). Similar to the classical Newton–Euler method, all the equations are written in body coordinate.

The spatial equation of motion states that the net force acting on a rigid body equals its rate of change of momentum [44]:

$$f_i^B = \frac{d}{dt}(I_i v_i) = I_i a_i + v_i \times^* I_i v_i \tag{22}$$

in this equation, $f_i^B$ represents the net force applied to $i$. The left side of this equation is the rate of change of body $i$'s angular momentum. $\times^*$, $v_i$, $a_i$ and $I_i$ are spatial force product sign, velocity, acceleration and moment of inertia, respectively. If the external force $f_i^x$ is applied to body $i$ and $f_j$ is applied by body $i$ to its successor body (body $j$), the following equation is written for body $i$.

$$f_i^B = f_i + f_i^x - \sum_{j \in \mu(i)} f_j \tag{23}$$

This equation is then reordered to obtain the recurrence relation of forces acting on body $i$:

$$f_i = f_i^B - f_i^x + \sum_{j \in \mu(i)} f_j \tag{24}$$

Equation (7) is solved by non-recursive algorithms, possessing a high computational complexity as $O(n^4)$. One the other hand, Recursive algorithms are much more efficient, possessing computational complexities of O(n) [42]. Hence, though non-recursive methods give preferable physical understanding, recursive algorithms are much better for implementation. Therefore, human and RoboWalk are modelled by RNEA and, for the sake of validation, the obtained results are compared with those of Newton's method in subsequent sections.

## Structural Optimization

As mentioned in previous sections, each leg of RoboWalk assists the user by applying two forces towards the user CoG. The resultant of these two forces are two forces along X and Z direction in sagittal plane. Since RoboWalk has one actuator on each leg, it is an under-actuated system. As a result of this matter, one cannot possess control on setting both forces to the desired value. Hence, by assuming the control strategy mentioned in equation (20), the assisting force in the Z direction is set but the creation of a force in X direction is inevitable. The effect of applying this horizontal force can be explained in such a way that the person is being pushed or pulled during walking and this might cause the user to fall.

This undesirable disturbing horizontal force could be vanished to some extent if the upper link of the robot could stay horizontal during the entire gait. In this way, most of the assisting force created by motors would remain in Z direction. For this purpose, either upper or lower link of the robot must have variable length. This could be done using another actuator or passive elements like springs for changing the robot link length. Many designs were checked and studied but they either increased the weight of the system or its complexity with no guarantee of the disturbing force reduction in the entire gait.

Another approach for decreasing this horizontal force is optimizing the dimensions of the robot such that the horizontal force is taken care of to some extent and low torque is required for RoboWalk motor. Trying to keep the motor torques as low as possible has two benefits. It allows using smaller and low power actuator system (motors and gears) and therefore, helps keeping robot's weight and manufacturing cost as low as possible. Hence, it is concluded that the cost function of the optimization problem should be as follows.

$$J = \sum {F_x}^2 + \sum {T_m}^2 \tag{25}$$

where $F_x$ and $T_m$ represent horizontal force and motor torque, respectively. In this optimization problem, the design parameters are chosen to be $L$ (distance between user CoG and RoboWalk knee), $r$ (lower link length), $R$ (arc radius) and $\theta_{opt}$(angle between $L$ and the line passing through human CoG and the roller that applies force to the user) which are all circled in Figure 8. The rest of the parameters are considered to be a function of $LrR\theta_{opt}T$parameter-set but the mass of each limb is assumed to remain constant during optimization.

Three methods were examined to design RoboWalk optimally. Since this problem is a continuous optimization problem, the Particle Swarm Optimization (PSO) method is used to obtain the optimal parameter values for each robot leg. It is worth mentioning that this optimization process is a constrained optimization problem and all dimensional parameters are bounded, giving the set of inequality constraints as follows:

$$\lambda_i^{\min} \leq \lambda_i \leq \lambda_i^{\max} \tag{26}$$

when finding optimal parameters, the geometrical and velocity constraints be violated. Some of these geometrical constraints are mentioned as.

$$x_0 + L\sin\theta_1 + r\sin(\theta_1 + \theta_2) = x_e \tag{27}$$

$$z_0 - L\cos\theta_1 - r\cos(\theta_1 + \theta_2) = z_e \tag{28}$$

$$\frac{\sin\theta_{opt}}{LL_1} = \frac{\sin\theta_{LL1}}{R} \tag{29}$$

$$\frac{\sin(\theta_{opt} + \theta_R)}{LL_2} = \frac{\sin\theta_{LL2}}{R} \tag{30}$$

### First approach for structural optimization

In the first approach, optimization is conducted for each gait sample separately. In other words, when the user steps a single sample time forward, valid combinations (population) of parameter-set values are chosen and the cost function is evaluated for each parameter-set. Consecutive parameter-sets are selected such that the cost function reduces in every step. In this approach, the cost function is proposed as

$$\begin{aligned} J_1 = w_1 \sum {F_x}^2 + w_2 \sum {T_m}^2 \\ + w_i \times Geometrical\, constraints \end{aligned} \tag{31}$$

Where $w_1$, $w_2$ and $w_i$ are the weights and $F_x$ and $T_m$ represent horizontal force exerted to the user and RoboWalk motor torque, respectively. The horizontal force is obtained according to user posture and RoboWalk configuration for the intended parameter-set. The structure of the optimization problem in the first approach is illustrated in Figure 9. As shown in the figure, the optimal parameters that minimize the cost function are obtained for every sample gait data ($k^{th}$ gait sample).

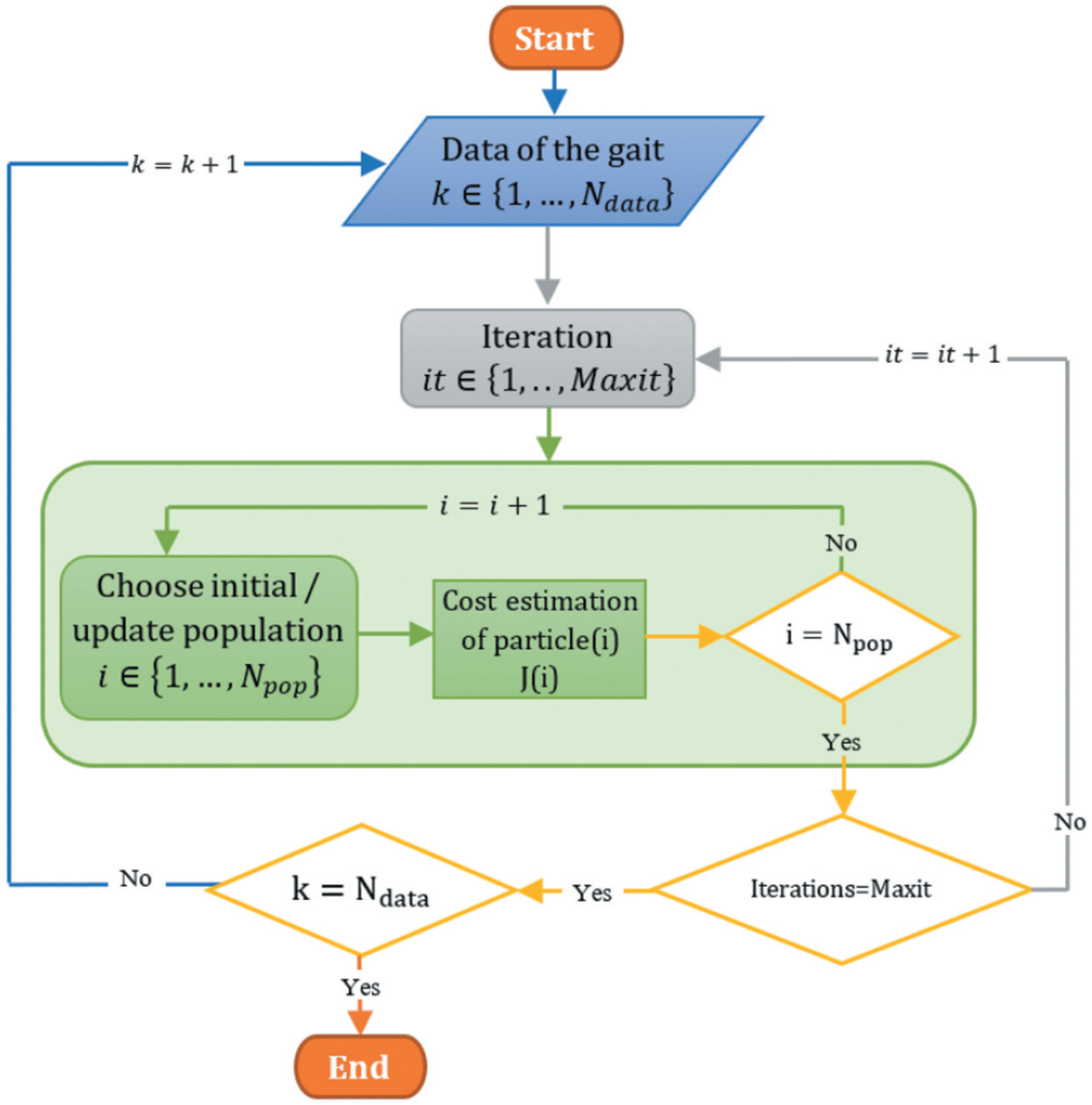

Figure 9. The first approach of optimization structure.

In this figure, $N_{data}$ is the number of data samples, $N_{pop}$ is particle population (number of parameter-sets) and *Maxit* is the number of iterations made for every sample to find the optimal structure for that gait sample. The pseudo-code of this structure is as follows.

```
Begin
  for i=1: all_samples
  for j= 1: iteration_number
  Update the population position
  if geometrical parameters are Complex;
  Cost(j)= inf;
  elseif geometrical parameters are Real;
  Calculation of F1, F2 and Tm in SSP;
  Checking geometrical constraints;
  Calculation of Cost;
  end if
  end for
  Find the Global_Best(i)
  end for
```

In this strategy, the Global_Best(i) is the best dimension set that minimizes the cost function in that particular sample. Hence, although this method might not have a unique answer, if a parameter-set causes the cost function to be in its minimum in most of the gait, those parameters are selected as the best configuration.

### *Second approach for structural optimization*

Most of the optimizations in the literature are performed by a strategy similar to this approach [26,32,33,35]. In this method, unlike the previous method that the optimization was done for every gait sample, the optimization is done for the entire gait at once. The structure of this optimization approach is depicted in Figure 10.

In this structure, unlike the first approach that consecutive parameter-sets were chosen for every gait sample, a parameter-set is chosen in every iteration of the optimization and the cost of choosing that parameter-set is estimated for the entire gait. Then, in the next iteration, another parameter-set is selected and the cost estimation is performed once again. The PSO algorithm that minimizes this cost function is capable of finding optimal structural parameters to minimize a cost over the entire gait cycle. The pseudo-code of this structure is demonstrated as follows.

```
Start
Initialize population randomly
Evaluate the initial population
Calculation of Cost;
for i= 1: iteration_number
for all of the population
Update the population position
all_samples= gait_data;
for j=1: all_samples
if geometrical parameters are Complex;
Cost(j)= inf;
elseif geometrical parameters are Real;
Calculation of F1, F2 and Tm in SSP;
```

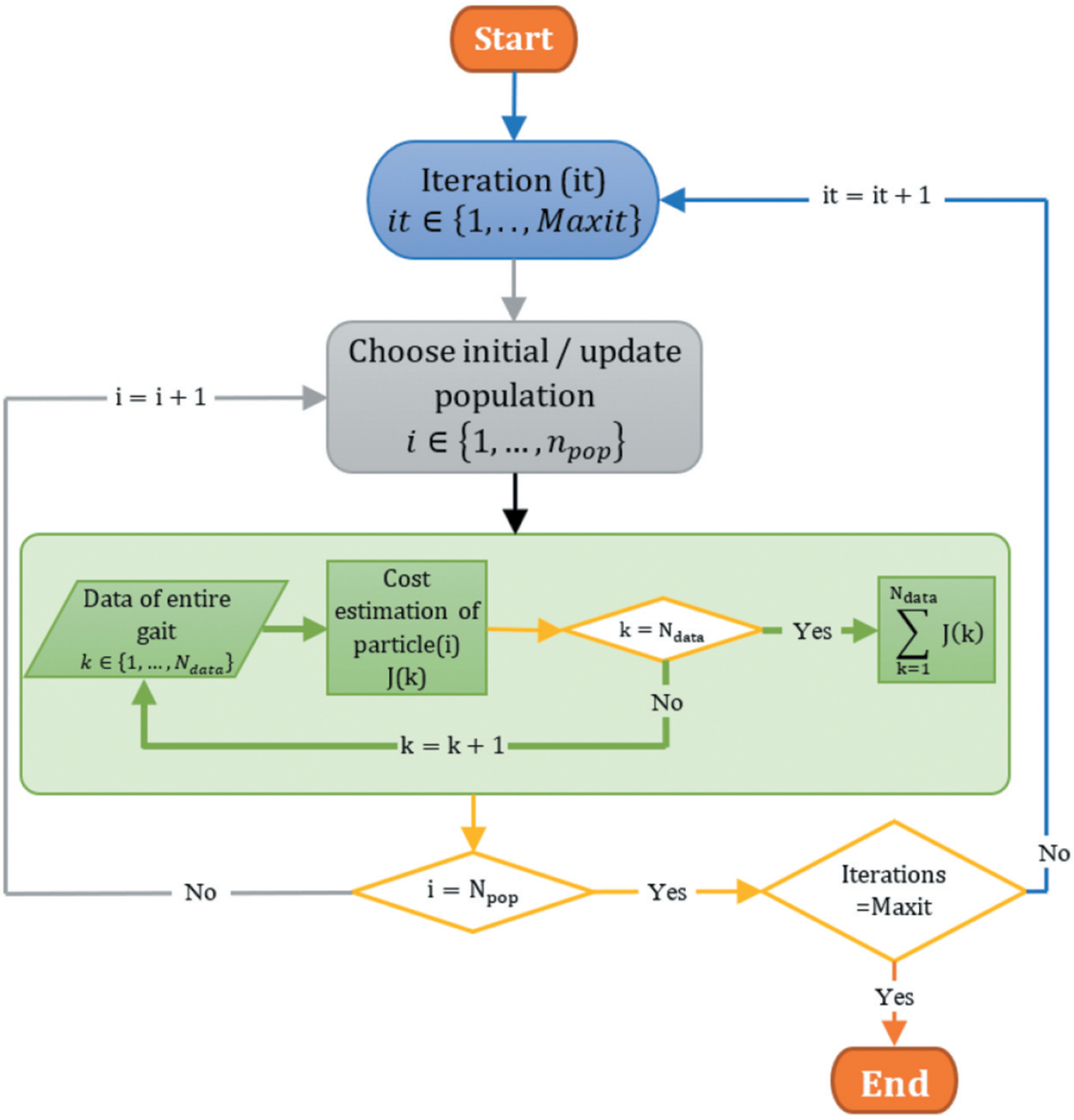

**Figure 10.** The second approach of optimization structure.

Checking geometrical constraints;
Calculation of Cost(j);
**end if**
**end for**
Overall_Cost= ∑Cost(j);
**end for**
**end for**

The cost function in this approach is similar to the previous.

$$J_2\big|_{gait\ sample} = w_1 {F_x}^2 + w_2 {T_m}^2 + w_i \times Geometrical\ constraints \tag{32}$$

This cost function should be estimated for every set of parameters for the entire gait (every iteration). Hence, it must be accumulated or averaged for each iteration as follow

$$J_2\Big|_{\substack{iteration \\ (gait)}} = \sum_{i=1}^{N_{data}} \left\{ \begin{array}{l} w_1 {F_x}^2 + w_2 {T_m}^2 \\ +w_i \times Geometrical\ constraints \end{array} \right\} \tag{33}$$

The only objection to this approach is that although it considers the horizontal force, the main contributing factor to the increase of knee-joint torque, it does not involve the joint torque directly.

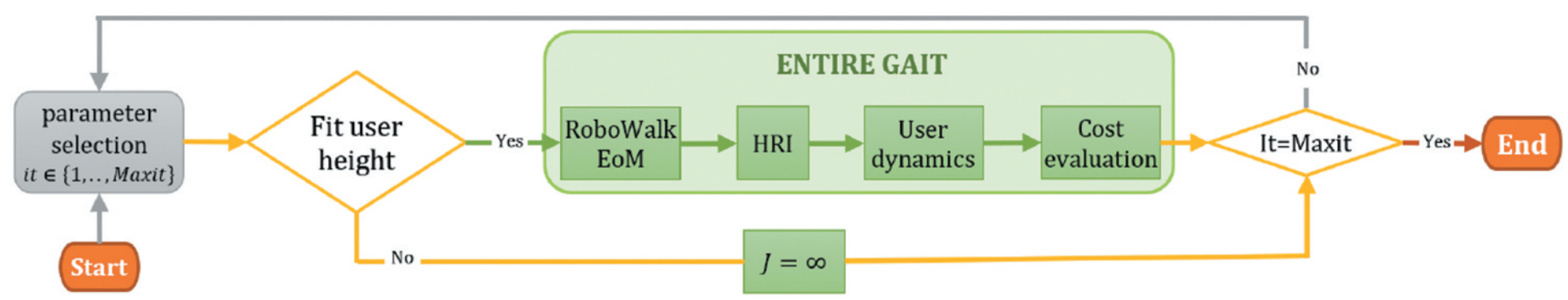

**Figure 11.** The optimization structure of the human-model-in-the-loop method.

## ***Human-model-in-the-loop method***

The human model-in-the-loop method is proposed to overcome the problem of the previous method. In this method, the human skeletal/musculoskeletal dynamics model is involved in each cost function evaluation. The structure of this method is illustrated in Figure 11 and explained in the subsequent concise form pseudo-code (the initialization is omitted).

**Start (Main Loop)**
**for** i= 1: iteration_number
**for** all of the population
Obtain population position
all_samples= gait_data;
**for** j=1: all_samples
**if** geometrical parameters are Complex or
the robot doesn't fit the user height;
Cost(j)= inf;
**elseif**
Find RoboWalk kinematics for parameters;
Obtain RoboWalk EoM and HRI;
Estimation of HRI effect on human model;
Calculation of Cost(j);
**end if**
**end for**
Overall_Cost= ∑Cost(j);
**end for**
**end for**

In the first step of this optimization method, random parameters are selected for the robot. These parameters are examined to fit for a specific user height. If the parameters selected could not fit the desired user, another set of parameters will be chosen. The valid set of parameters will be used for the robot in the entire gait. By knowing the user's forward kinematics, RoboWalk inverse kinematics will be estimated using the selected parameters. Knowing the kinematics leads to obtaining the forces applied to the user by the robot. The interaction between robot and user in every iteration causes different assisting forces and different knee torque. Hence, the cost function to obtain the least knee torque and actuator torque for the user is as follows.

$$J_3 = \sum_{i=1}^{N_{data}} \left\{ \begin{array}{l} w_1 {\tau_{knee}}^2 + w_2 {T_{m_L}}^2 + w_3 {T_{m_R}}^2 \\ + w_i \times geometrical\ constraints \end{array} \right\} \tag{34}$$

In this cost function, $\tau_{knee}$, $T_{m_L}$ and $T_{m_R}$ are the knee torque for each user's leg, actuator torque of the left and right leg of RoboWalk, respectively. This cost function should be estimated for every set of parameters (every iteration). Hence, it must be added or averaged for each iteration (entire gait).

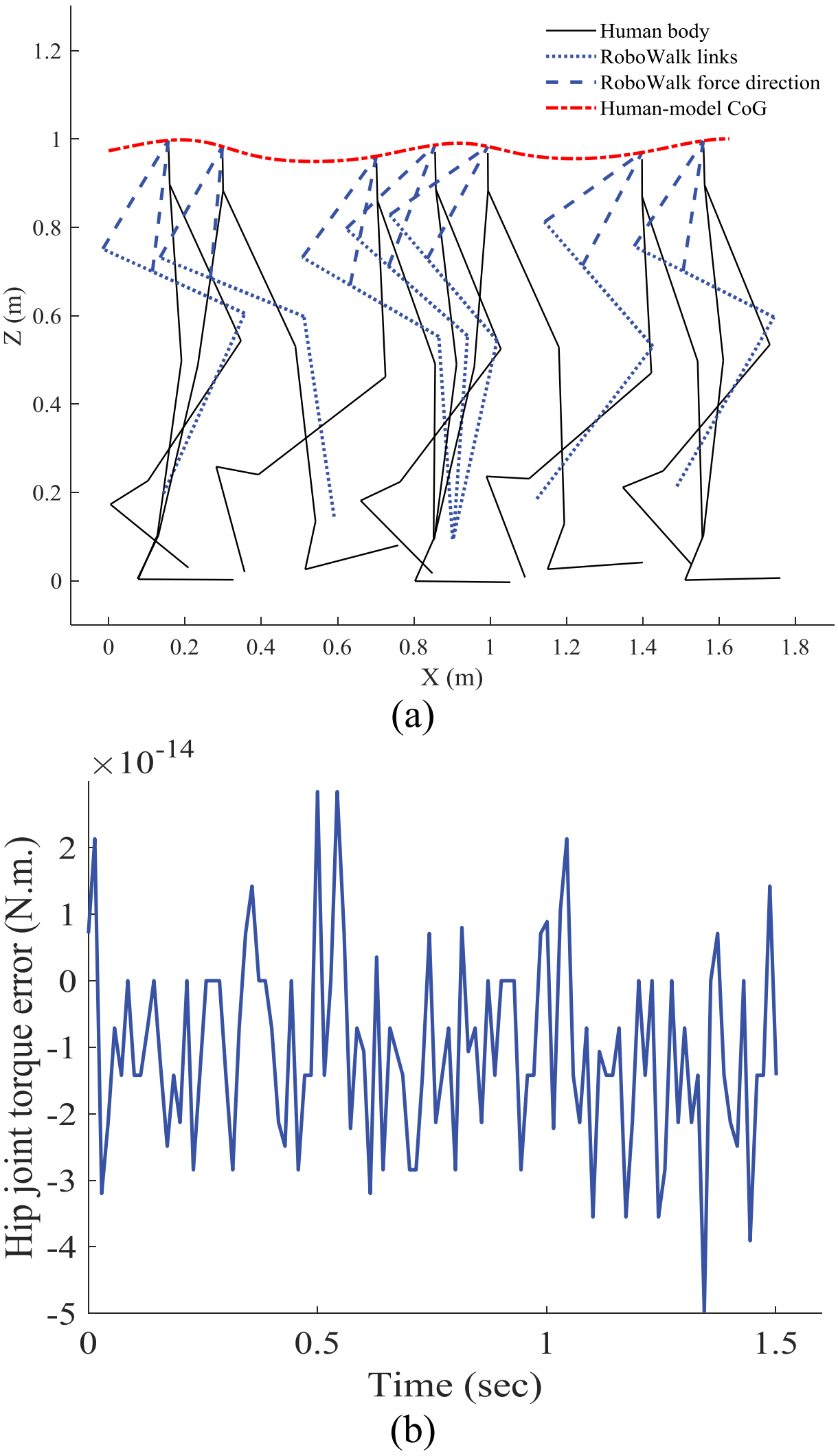

**Figure 12.** A) Kinematics verification and b) dynamics verification (error between Newton algorithm and RNEA).

## Obtained Results and Discussion

The simulations in this research are carried out on Windows 10, core i7, CPU (2.8 GHz), memory 16 GB. The human mass and geometrical properties were extracted from the human skeletal model in Opensim software. The human kinematics (mentioned in equation (1)) consisted of 188 samples for the kinematics verification and 54 samples for the optimization. The kinematics of the augmented system is verified in Figure 12.a. In this figure the dashed-dotted line is the locus of human model CoG, the dashed line is the direction of assisting forces applied to the user (directed towards user CoG), dotted lines are RoboWalk left leg expansion and contraction mechanism and solid lines are the lower extremities of human user. In order to verify the models, the Newton method is compared to the numerical RNEA. For arbitrary trajectories of the joints, joint forces are obtained by these methods and compared to each other. In Figure 12.b the result of both methods is compared.

As it can be observed in this figure, both methods yield the same results and the error between them (of the order of $10^{-14}$) is due to computation round-off.

The robot's dimensional specifications are extracted from Honda's bodyweight support system [45]. Since our robot has more sensors and electronic boards for research purposes and a heavier seat, we decided to assume the weight to be 8 Kg in a conservative estimation, which is a typical weight for non-anthropomorphic assistive devices [46,47] without backpacks. Tables 1 and 2 illustrate human and robot's mass properties and geometrical specifications.

The geometric properties of human and RoboWalk is as follows.

Using these geometrical and mass specifications and obtaining the kinematic gait data from Opensim software, the RoboWalk influence on user's knee force are estimated and shown in Figure 13. The solid and dotted-dashed line is after and before using RoboWalk, respectively.

In this figure, $F_{KR}$and $F_{KL}$represent the joint force of right and left knee, respectively. Elders are most vulnerable at knee joints due to weak muscles and arthritis. In patients suffering from knee arthritis, the force exerted on knee causes pain. It is demonstrated in this figure that the joint force (bone on bone forces) has decreased dramatically after utilizing RoboWalk. Simulations showed an increase in joint torques after using the device in most of the gait cycle which is mainly caused by the disturbing horizontal force. This is an important fact that could harass, exhaust, or lead to the fall of the user. Consequently, most of the force is tolerated by the muscles and the bone on bone forces are reduced considerably. In order to decrease the disturbing force, we can either reduce the percentage of RoboWalk assisting force or change the structure. Accordingly, a structural optimization to minimize user knee joint torque and robot motor torque is conducted.

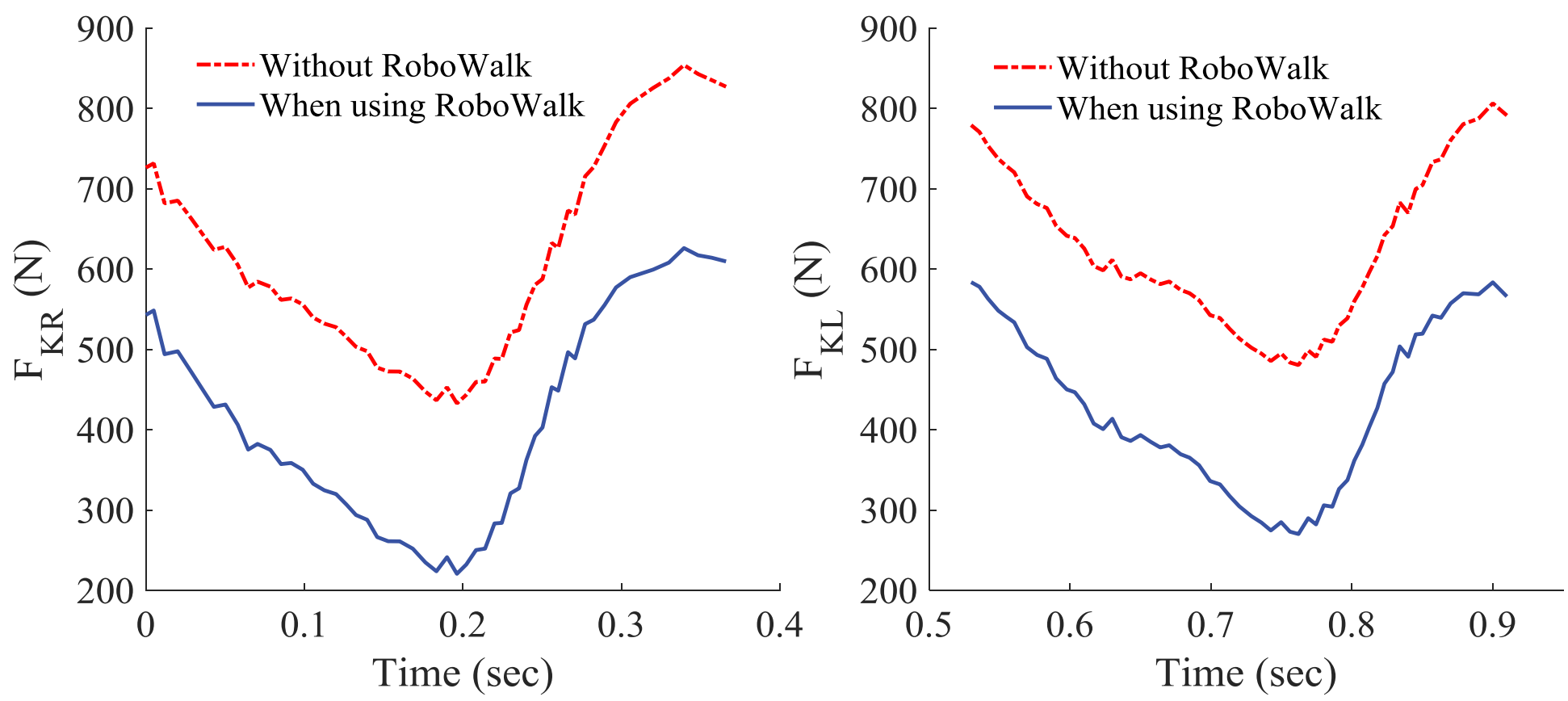

**Figure 13.** Right and left knee force with and without RoboWalk.

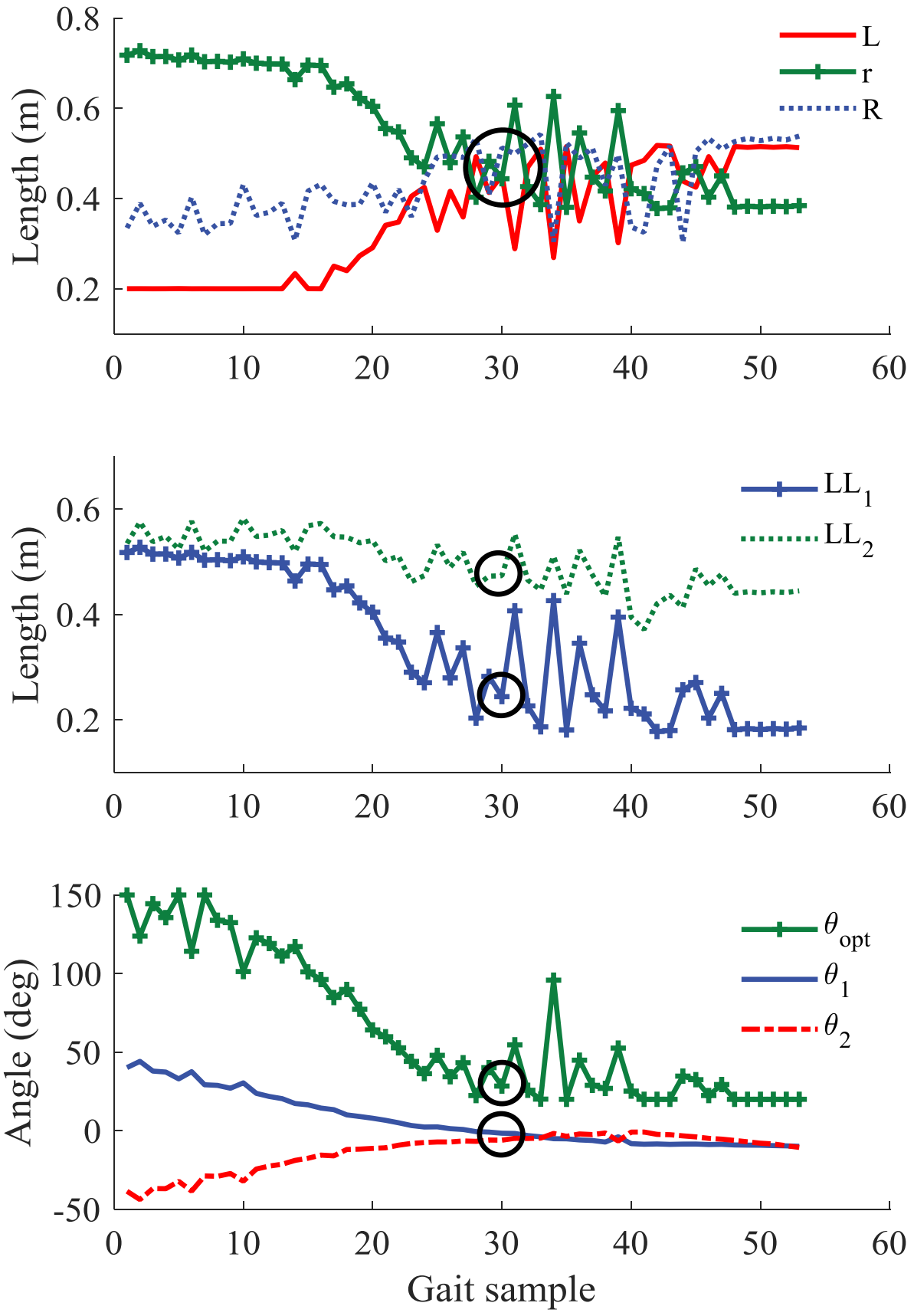

Figure 14. Optimized parameters obtained by the first approach.

### *First approach of design optimization*

The optimization of all the methods was performed by PSO with a population of 500 particles. Optimization was done for the left leg SSP when more data was collected than other SSP phases (54 data samples). The first approach achieved to the following results (Figure 14).

In this case, the force and robot motor torque are shown in Figure 15.

As shown in Figure 14 in the first half of SSP, the average values of all parameters are different from the other half's average values. Gait sample number 30 could be assumed as a suitable average sample for all parameters. Accordingly, good results for this optimization approach are summarized in Table 3.

This approach benefits of its low calculation time for solving the optimization problem but unfortunately, there is no precise optimal value for entire gait and an average value should be considered. There is no guarantee that the selected average value is an appropriate value for the rest of the gait.

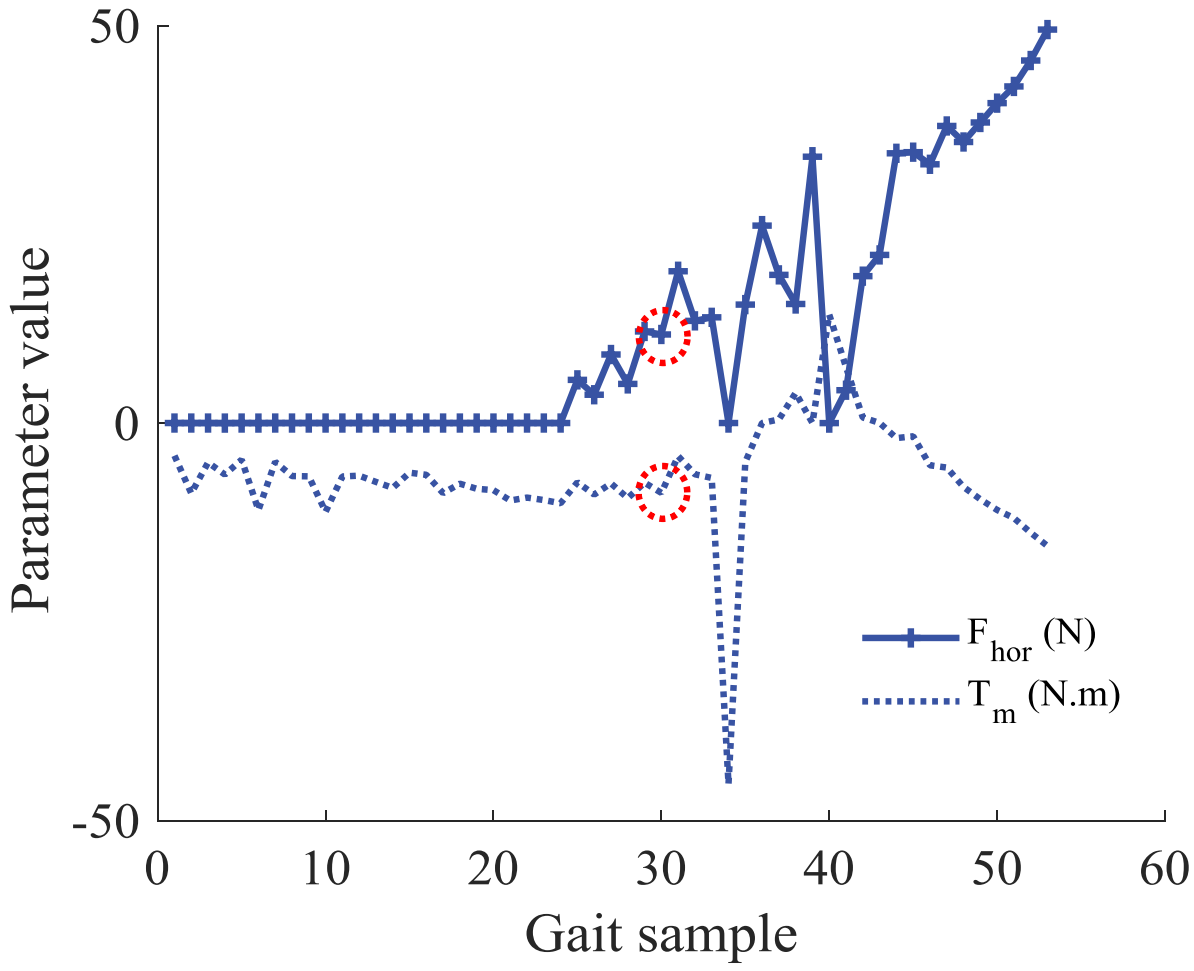

**Figure 15.** RoboWalk motor torque and the horizontal force applied to the user in the first approach.

### Second approach of design optimization

The optimization in this method is performed in the same situation as the previous; namely, the data samples and bounding of parameters are the same in both methods. The results of this optimization are illustrated in Figure 16.

In this case, the force and robot motor torque are demonstrated in Figure 17.

As it is shown in these figures, robot parameters have converged to specific values. These values are optimized throughout the entire SSP and are reliable to produce the least horizontal force by using the least motor torque value. The parameter values after convergence of the algorithm are specified in Table 4.

The calculation time of this approach is less than the previous and the parameters converge to the values that minimize the overall cost function. The only deficiency of this approach is that it only deals with actuator torque and horizontal force and it does not include the user knee-joint actuation torque in the cost function. This problem is addressed in the next approach. In addition, $LL_1$ = 11 cm means that the distance between the knee joint of RoboWalk and the seat is approximately 11 cm. Which could cause problems when the user intends to sit or crouch. Since there is no control on $LL_1$ and $LL_2$ and geometrical constraints attain them, this could be considered as another drawback of this approach.

### Human-model-in-the-loop method

This method is suggested to solve the problems and deficiencies of previous methods. In this method, actuator torque and knee-joint torque of the human model is involved. Results of dimensional optimization for this approach is demonstrated in Figure 18.

The parameter values after convergence of the algorithm is specified in Table 5.

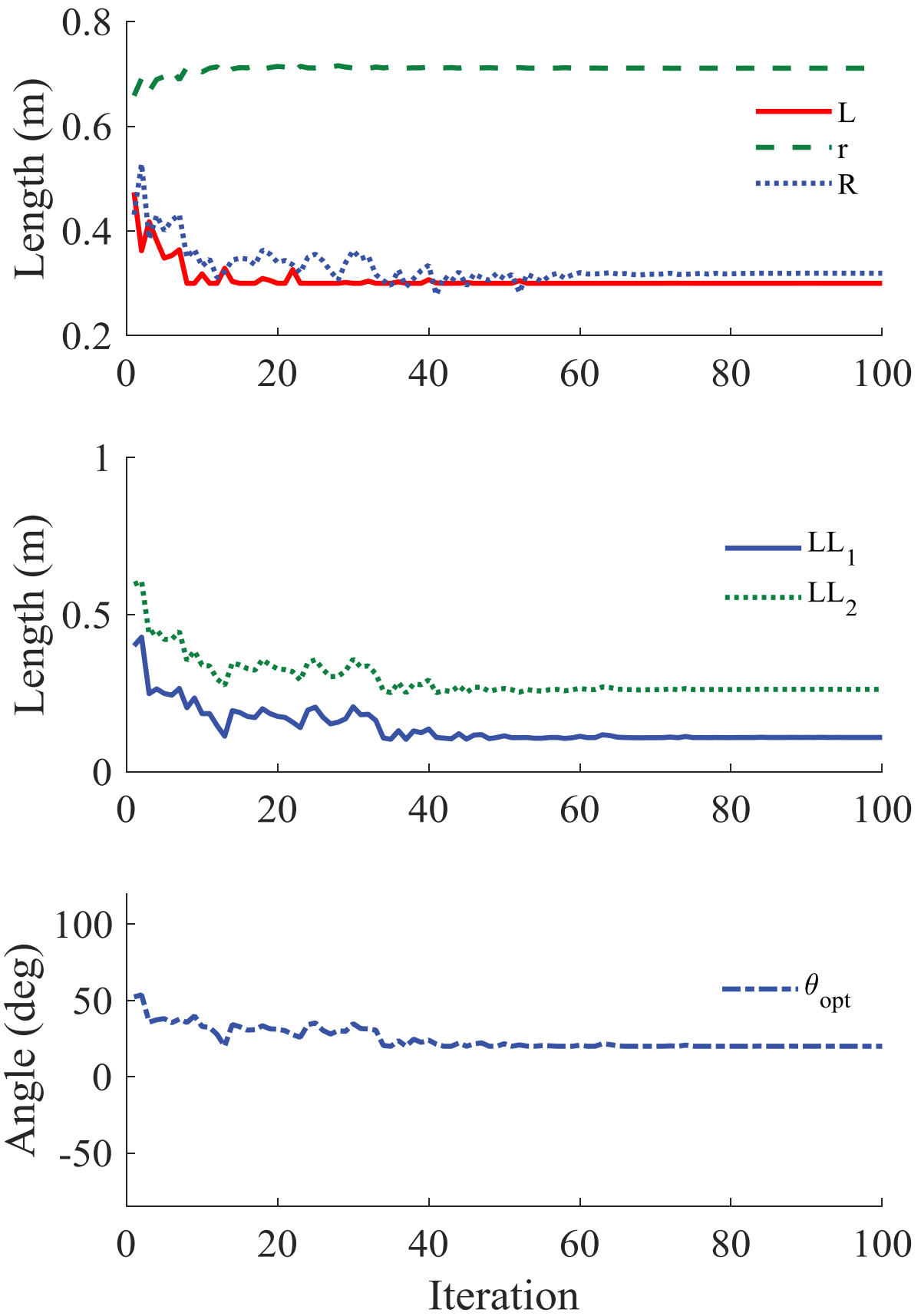

**Figure 16.** Optimized parameters obtained by the second approach.

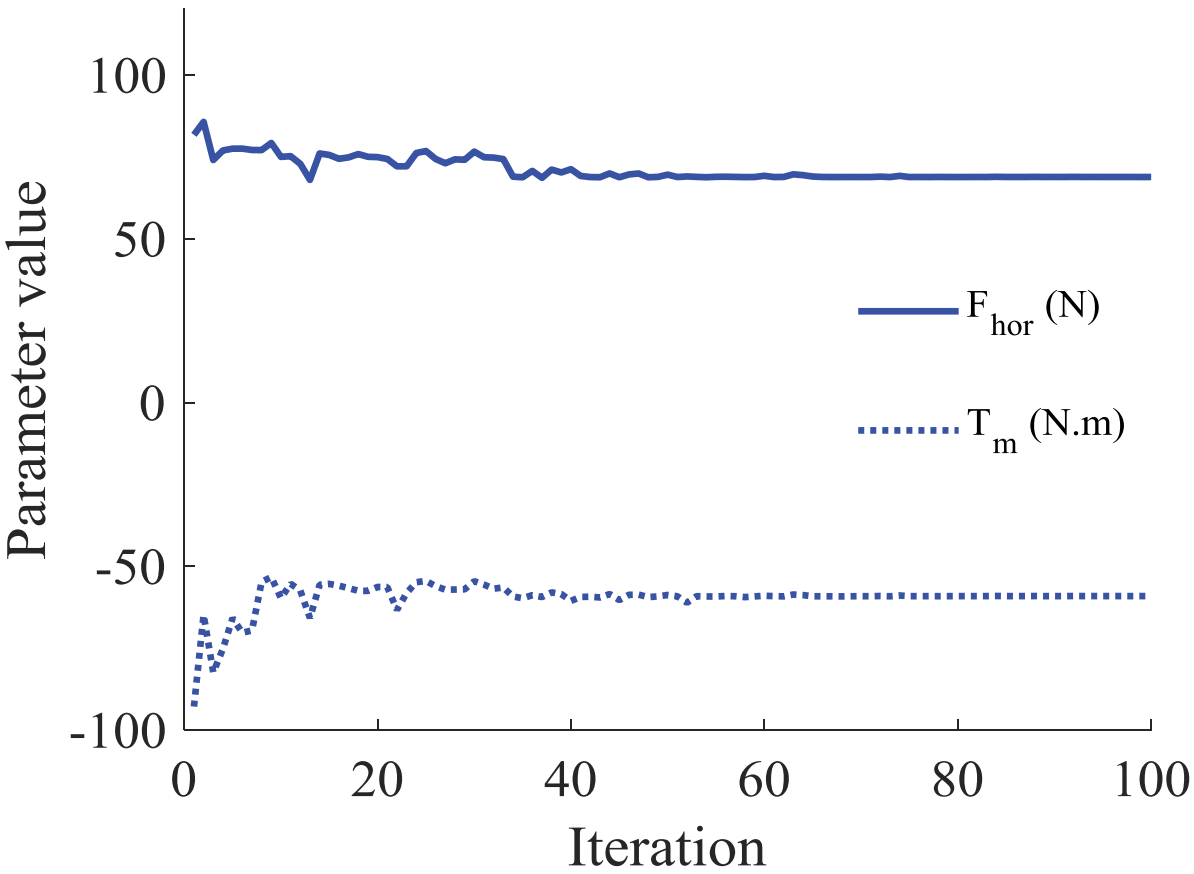

**Figure 17.** RoboWalk motor torque and horizontal force applied to the user in the second approach.

Using these optimal parameters, user knee-joint torque is obtained and illustrated in Figure 19.

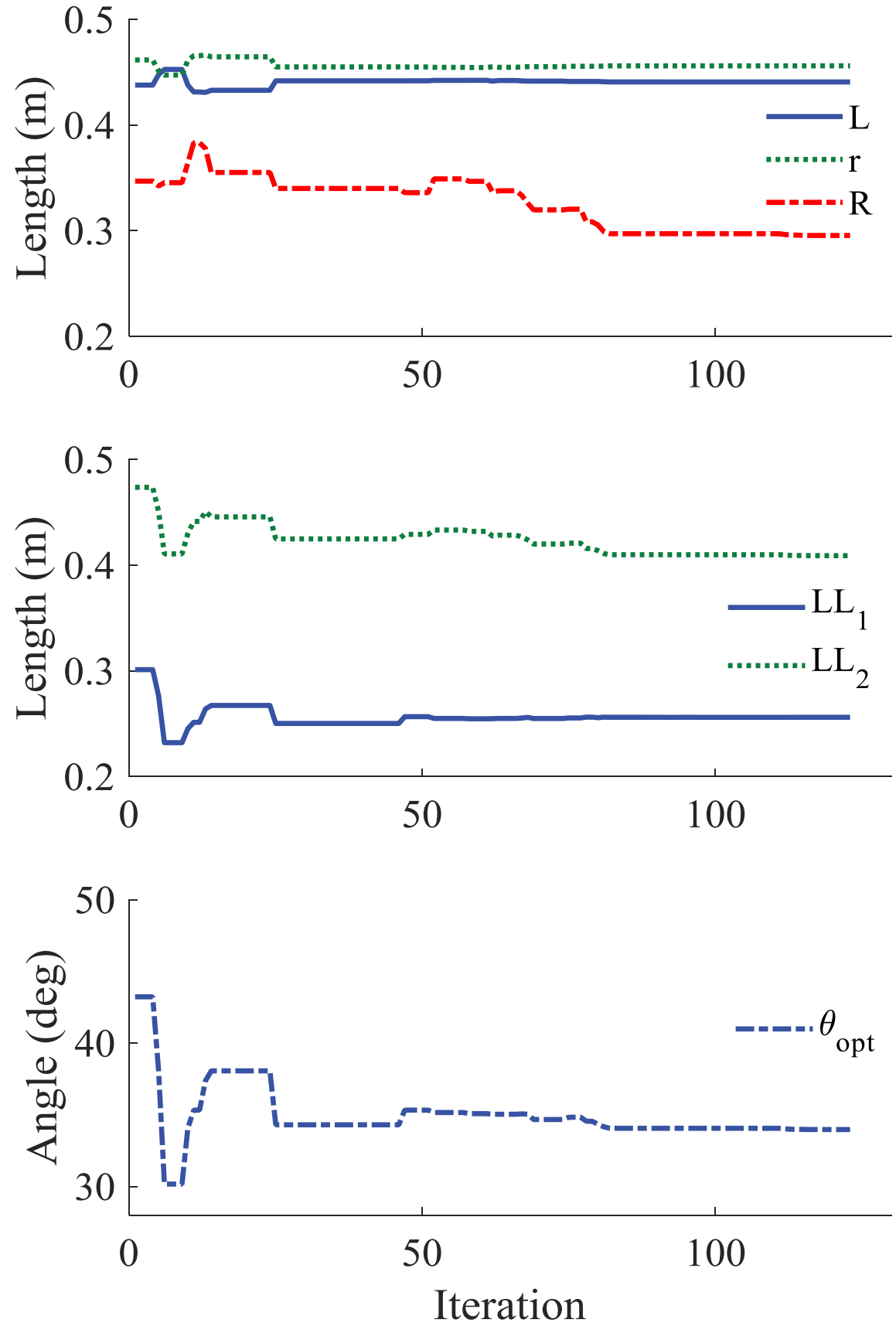

**Figure 18.** Optimized parameters obtained by the human-model-in-the-loop method.

In this figure, the solid line is the joint torque obtained after using the optimal values, the dashed line is the user joint torque obtained after using the intuitive values (non-optimal joint torque) and the dotted line is human joint torque before utilizing RoboWalk. The hatched part of the plot is the area that the user's joint torque is improved by using optimal values. Figure 20 depicts the speed-torque diagram of RoboWalk actuator after optimization is performed by each method.

In the previous design, RoboWalk actuators should have provided 70 N.m. nominal torque in approximately 6rpm velocity. After optimizing the device design by the first, second and HMIL approaches, the required torque of 65, 55 and 25 N.m. were achieved in approximately 6,7 and 20 rpm, respectively. Therefore, the HMIL approach is superior to other mentioned methods in the sense of obtaining the best actuator system. As a result of this valuable achievement, this design's required motor and gearbox are lighter, more likely to be back-drivable and even more affordable in price. This means that the 630 grams EC 90 flat Ø90 mm, brushless, 360 W, with Hall sensors that deliver a nominal torque of 953mNm in nominal speed of 2340rpm and a 30:1 gearbox is a very conservative and suitable selection for RoboWalk motor. This selection is far from good since it is even lighter than what we expected and used in our simulations for RoboWalk upper link mass (Table 1).

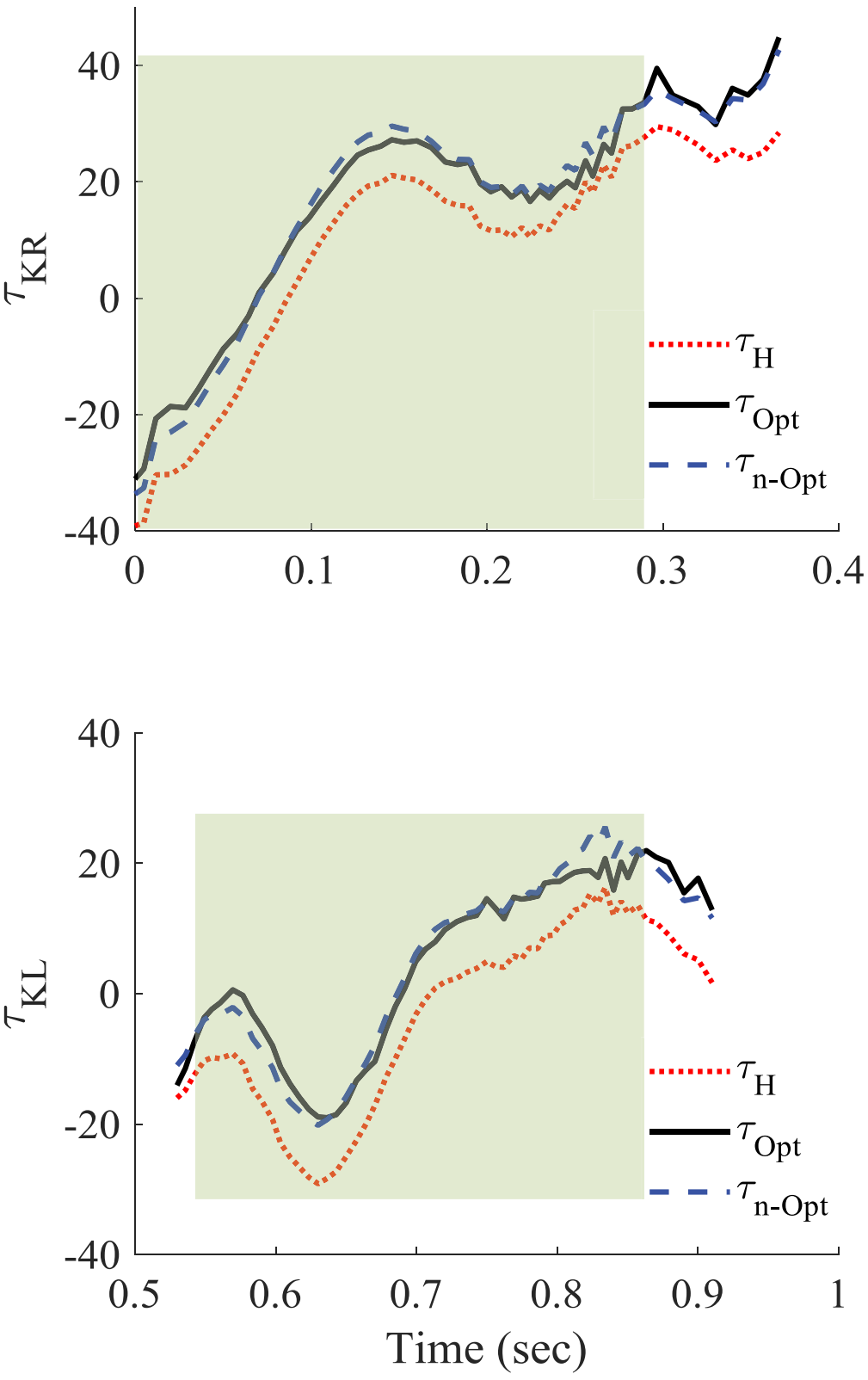

**Figure 19.** The user knee-joint torque with and without RoboWalk (before and after optimization).

As stated before, if a low torque (0–8 N.m.) is applied by the robot knee actuator, the seat of the robot remains attached to the user buttocks (Figure 21).

The results of the optimization showed that the least actuator torque was obtained by the HMIL approach and the amount of torque, in that case, was at least 15 N.m.; therefore, since the 8 N.m. actuator torque is never reached, the robot seat does not get detached from the user pelvis and assuming a perfect connection between them was reasonable in the modelling.

## Conclusions and Discussion

In this research, a lower-limb assistive exoskeleton, named RoboWalk, was presented and its operational strategy was explained in detail. After modelling the augmented human-robot system in 2D (sagittal plane) and verifying the kinematics and dynamics of this multi-body system, a simple control strategy was applied and RoboWalk performance and its effect on the user were investigated. The results showed that the robot reduced the felt bodyweight of the user by 33% using a 70 N.m. motor in approximately 6 rpm angular velocity but applied a large unwanted disturbing horizontal force to the user. The disturbing force causes the user to feel uncomfortable by increasing knee joint torque and

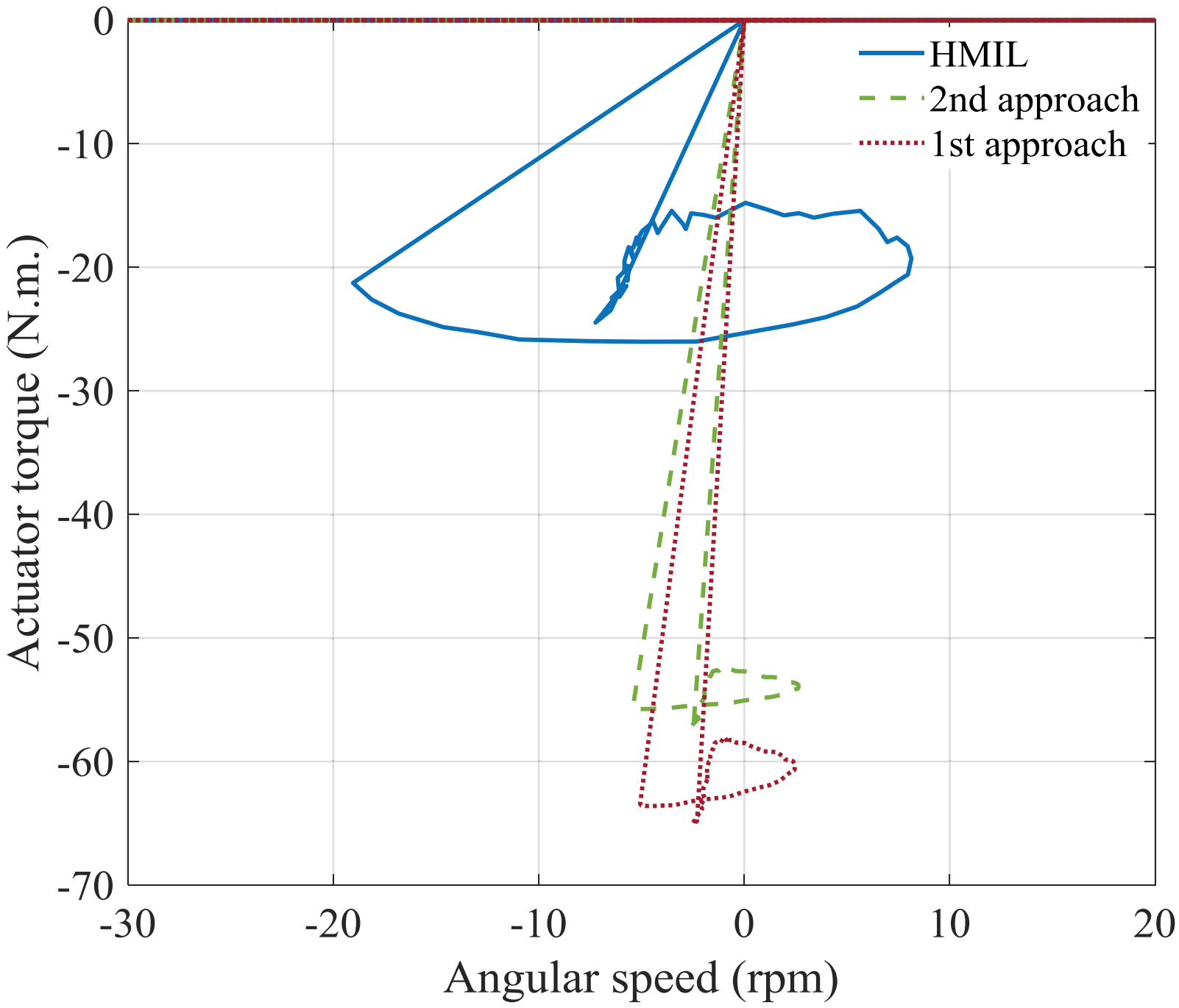

**Figure 20.** Speed-torque diagram of the left leg actuators before and after optimization in SSP (the torques in DSP are set to zero).

**Table 1.** Human and RoboWalk mass properties.

| Link | Mass (Kg) |
|---|---|
| Foot | 1.56 |
| Shank | 3.7 |
| Thigh | 9.3 |
| Pelvis | 11.78 |
| Trunk | 34.24 |
| Approximate robot upper link mass | 3 |
| Approximate robot lower link mass | 1 |
| Human total weight | 75.14 |
| Total robot weight | 8 |

**Table 2.** Human and RoboWalk geometrical properties.

| Link | Length (cm) |
|---|---|
| Foot | 25 |
| Ankle joint height from the ground | 10 |
| Shank | 40 |
| Thigh | 40 |
| Pelvis | 17 |
| Trunk | 60 |
| Robot upper link | 55 |
| Robot lower link | 55 |
| Horizontal distance of robot ankle with respect to human ankle | 5 |
| Vertical distance of robot ankle with respect to human ankle | 0 |
| Circle arc radius | 20 |

**Table 3.** Parameters obtained after optimization by the first approach.

| parameter | L | r | R | $LL_1$ | $LL_2$ |
|---|---|---|---|---|---|
| **Average value** | 0.45 | 0.45 | 0.51 | 0.24 | 0.47 |
| parameter | $\theta_{opt}$ | $\theta_1$ | $\theta_2$ | $F_{hor}$ | $T_m$ |
| **Average value** | 28.4 | −1.6 | −6 | 11.1 | −8.8 |

**Table 4.** Parameters obtained after optimization by the second approach.

| parameter | L | r | R | $LL_1$ |
|---|---|---|---|---|
| **Average value** | 0.3 | 0.71 | 0.32 | 0.11 |
| parameter | $LL_2$ | $\theta_{opt}$ | $F_{hor}$ | $T_m$ |
| **Average value** | 0.26 | 20 | 69 | −59 |

**Table 5.** Parameters obtained after optimization by the human-in-the-loop approach.

| parameter | L | r | R | $LL_1$ | $LL_2$ | $\theta_{opt}$ |
|---|---|---|---|---|---|---|
| **Average value** | 0.44 | 0.46 | 0.3 | 0.26 | 0.4 | 34 |

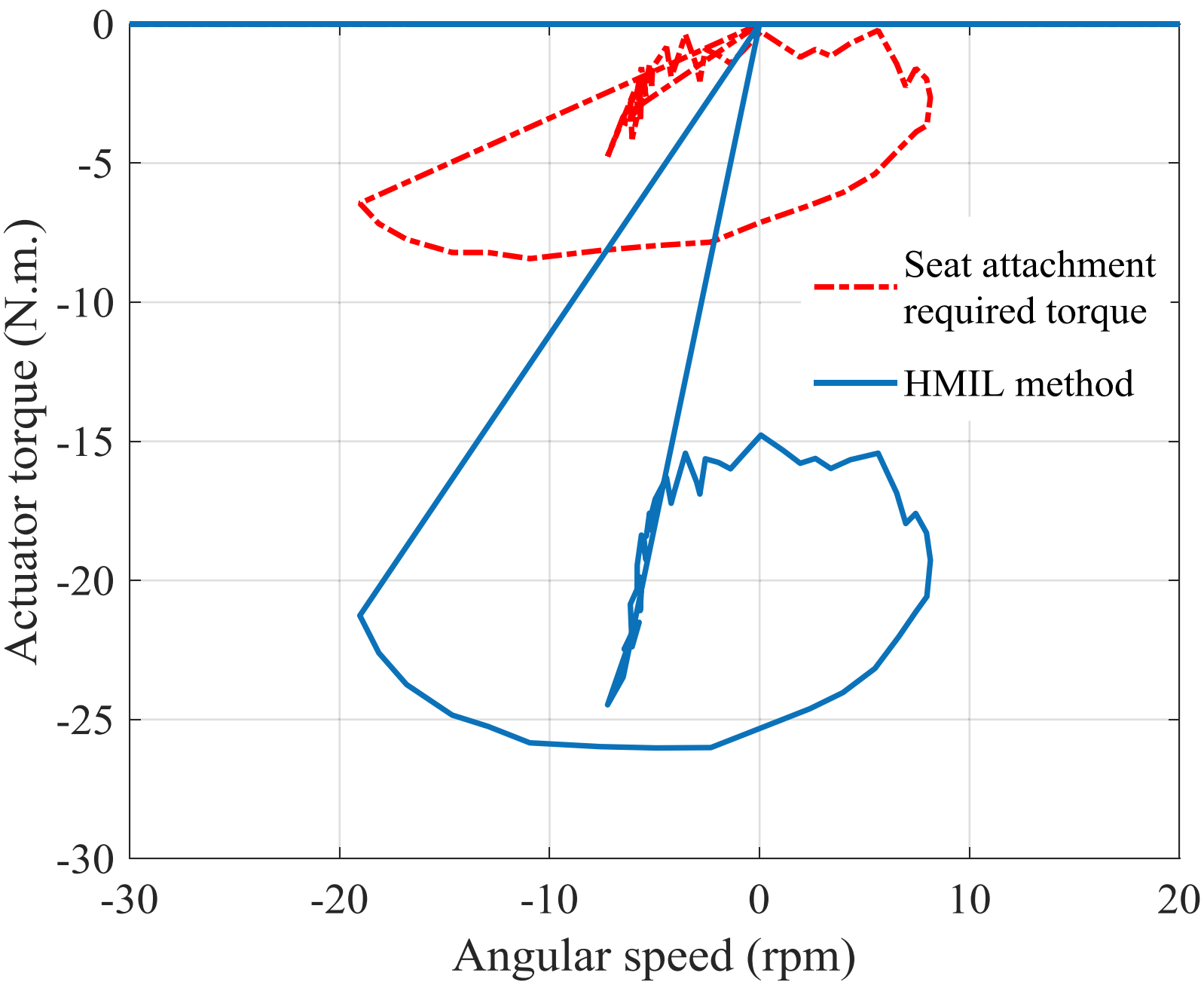

**Figure 21.** Comparing the actuator torque required for maintaining the seat attached to the pelvis and the minimum torque required for assisting the user (obtained by the HMIL approach).

might eventually lead to his/her fall. In order to overcome this problem, dimensional optimization was performed using different strategies. The optimization problems were solved by PSO algorithm and the masses of the limbs were considered as constant values during optimization. In the first method best robot parameters were found for every sample of the gait. Therefore, there is no convergence to any particular parameter-set for the entire gait and an average parameter-set suitable for most of the gait should be chosen as the optimal set. In the second method, every parameter-set was tested for the entire gait to minimize the disturbing force and actuator torque. The significance of the HMIL optimization is to add the user joint torque by utilizing the human-robot augmented model. The performance of these strategies is summarized in Table 6.

**Table 6.** Comparison of different optimization methods.

| | First method | Second method | Human-model-in-the-loop method |
|---|---|---|---|
| Convergence | - | + | + |
| Human joint torque after optimization | High | Moderate | Low |
| RoboWalk actuator torque after optimization | High | Moderate | Low |
| Overall system weight | Moderate | Moderate | Low |
| Computational time | Low | Very Low | High |

The computational time required for the first, second, and HMIL methods were 1, 0.5, and 5 hours, respectively. On average, the HMIL method reduced the knee joint torque by approximately 2 N.m. more than the best result obtained by the other two approaches. In addition, as it is shown in Figure 20, the best actuator speed-torque diagram is obtained for the HMIL approach. Therefore, according to the simulations, the proposed HMIL optimization strategy was found to be more useful than other strategies. Two significant benefits of the HMIL optimization are that in addition to the fact that the user felt less disturbing joint torques, the motor's required specifications decreased such that not only the system gets lighter and cheaper, but it also remains in the region of back-drivability, which is essential for user safety. In addition, having a lightweight actuator leads to less disturbing forces applied to the leg in the swing phase. It is worth noting that this optimization strategy could be used for other exoskeletons or other human models, such as human models in Opensim or Anybody modelling systems.

In the future, we hope to add some passive elements to the robot to eliminate the remaining undesirable disturbing forces in different human gaits (e.g. walking, sit to stand movement, climbing stairs, etc.). Designing a controller for the stance leg instead of using a simple control strategy for applying a constant upward assistive force to the user is also an exciting future research topic. Also, activating the robot swing leg actuator is now in progress to recover balance after perturbations.

## Disclosure statement

No potential conflict of interest was reported by the authors.

## ORCID

Mahdi Nabipour http://orcid.org/0000-0002-6263-4469

## References

[1] S.A.A. Moosavian, M. Alghooneh, and A. Takhmar, *Cartesian approach for gait planning and control of biped robots on irregular surfaces*, I. J. Humanoid. Rob. 06 (4) (2009), pp. 675–697. doi:10.1142/S0219843609001942.

[2] M. Ezati, M. Khadiv, and S.A.A. Moosavian, *An investigation on the usefulness of employing a two-segment foot for traversing stairs*, I. J. Humanoid. Rob. 14 (2017), pp. 1–31.

[3] M. Khadiv, A. Herzog, S.A.A. Moosavian, and L. Righetti, *Walking control based on step timing adaptation*, IEEE. Trans. Rob. (3) 2020, 10.1109/TRO.2020.2982584

[4] I. Reichl, W. Auzinger, H.-B. Schmiedmayer, and E. Weinmüller, *Reconstructing the knee joint mechanism from kinematic data*, Math. Comput. Model. Dyn. Syst. 16 (5) (2010), pp. 403–415. doi:10.1080/13873954.2010.507094.

[5] M. Pietrusinski, I. Cajigas, G. Severini, P. Bonato, and C. Mavroidis, *Robotic gait rehabilitation trainer*, IEEE/ASME. Trans. Mechatron. 19 (2014), pp. 490–499.

[6] G. Colombo, M. Joerg, R. Schreier, and V. Dietz, *Treadmill training of paraplegic patients using a robotic orthosis*, J. Rehabil. Res. Dev. 37 (2000), pp. 693–700.

[7] S.K. Banala, S.H. Kim, S.K. Agrawal, and J.P. Scholz, *Robot assisted gait training with active leg exoskeleton (ALEX)*, IEEE. Trans. Neural. Syst. Rehabil. Eng. 17 (1) (2009), pp. 2–8. doi:10.1109/TNSRE.2008.2008280.

[8] H. Kawamoto, and Y. Sankai, *Power Assist System HAL-3 for Gait Disorder Person*, In International Conference on Computers for Handicapped Persons, Springer Berlin Heidelberg, (2002), pp. 196-203. doi:10.1007/3-540-45491-8_43.

[9] G. Chen, P. Qi, Z. Guo, and H. Yu, *Mechanical design and evaluation of a compact portable knee–ankle–foot robot for gait rehabilitation*, Mech. Mach. Theory. 103 (2016), pp. 51–64. doi:10.1016/j.mechmachtheory.2016.04.012.

[10] I. Díaz, J.J. Gil, and E. Sánchez, *Lower-limb robotic rehabilitation: Literature review and challenges*, J. Rob. 2011 (2011). doi:10.1155/2011/759764.

[11] A.B. Zoss, H. Kazerooni, and A. Chu, *Biomechanical design of the Berkeley lower extremity exoskeleton (BLEEX)*, IEEE/ASME. Trans. Mechatron. 11 (2) (2006), pp. 128–138. doi:10.1109/TMECH.2006.871087.

[12] J.E. Pratt, B.T. Krupp, and C.J. Morse, *The RoboKnee: An exoskeleton for enhancing strength and endurance during walking*, Proceedings - IEEE International Conference on Robotics and Automation New Orleans. LA, USA, (2004), doi:10.1109/ROBOT.2004.1307425.

[13] C.J. Walsh, K. Endo, and H. Herr, *A quasi-passive leg exoskeleton for load-carrying augmentation*, Int. J. Humanoid. Rob. 4 (3) (2007), pp. 20. doi:10.1142/S0219843607001126.

[14] W.V. Dijk and H.V.D. Kooij, *XPED2: A passive exoskeleton with artificial tendons*, IEEE Robot. Autom. Mag. 21 (4) (2014), pp. 56–61. doi:10.1109/MRA.2014.2360309.

[15] S.H. Collins, M.B. Wiggin, and G.S. Sawicki, *Reducing the energy cost of human walking using an unpowered exoskeleton*, Nature 522 (7555) (2015), pp. 212. doi:10.1038/nature14288.

[16] S. Krut, M. Benoit, E. Dombre, and F. Pierrot *MoonWalker, a Lower Limb Exoskeleton Able to Sustain Bodyweight Using a Passive Force Balancer*, 2010 IEEE International Conference on Robotics and Automation, (2010), pp. 2215-2220, doi: 10.1109/ROBOT.2010.5509961.

[17] Y. Ikeuchi, J. Ashihara, Y. Hiki, H. Kudoh, and T. Noda, *Walking assist device with bodyweight support system*, Proceedings of the 2009 IEEE/RSJ international conference on Intelligent robots and systems, St. Louis, MO, USA, 2009.

[18] W. Jun Ashihara, T. Yoshihisa Matsuoka, W. Hiroshi Matsuda, W. Yutaka Hiki, and W. Makoto Shishido, *Walking Assist Device*, HONDA MOTOR CO., LTD, Tokyo, Japan, 2015, pp. 1–41.

[19] *Design analysis of a passive weight-support lower-extremity-exoskeleton with compliant knee-joint*.

[20] Z. Lovrenovic, *Development and Testing of Passive Walking Assistive Exoskeleton with Upward Force Assist*, Doctoral dissertation, Université d'Ottawa/University of Ottawa, 2017.

[21] J. Zhang, P. Fiers, K.A. Witte, R.W. Jackson, K.L. Poggensee, C.G. Atkeson, and S.H. Collins, *Human-in-the-loop optimization of exoskeleton assistance during walking*, Science 356 (6344) (2017), pp. 1280–1284. doi:10.1126/science.aal5054.

[22] W. Felt, J.C. Selinger, J.M. Donelan, and C.D. Remy, *"Body-in-the-loop": Optimizing device parameters using measures of instantaneous energetic cost*, PloS One 10 (8) (2015), pp. e0135342. doi:10.1371/journal.pone.0135342.

[23] M. Kim, Y. Ding, P. Malcolm, J. Speeckaert, C.J. Siviy, C.J. Walsh, and S. Kuindersma, *Human-in-the-loop Bayesian optimization of wearable device parameters*, PloS One 12 (9) (2017), pp. e0184054. doi:10.1371/journal.pone.0184054.

[24] K. Tanghe, E. Aertbeliën, J. Vantilt, M. Moltedo, T. Bacek, D. Lefeber, and J. De Schutter, *Realtime delayless estimation of derivatives of noisy sensor signals for quasi-cyclic motions with application to joint acceleration estimation on an exoskeleton*, IEEE. Rob. Autom. Lett. 3 (3) (2018), pp. 1647–1654. doi:10.1109/LRA.2018.2801473.

[25] K. Tanghe, F. De Groote, D. Lefeber, J. De Schutter, and E. Aertbeliën, *Gait trajectory and event prediction from state estimation for exoskeletons during gait*, IEEE. Trans. Neural. Syst. Rehabil. Eng. 28 (1) (2019), pp. 211–220. doi:10.1109/TNSRE.2019.2950309.

[26] J. Ortiz, T. Poliero, G. Cairoli, E. Graf, and D.G. Caldwell, *Energy efficiency analysis and design optimization of an actuation system in a soft modular lower limb exoskeleton*, IEEE. Rob. Autom. Lett. 3 (1) (2018), pp. 484–491. doi:10.1109/LRA.2017.2768119.

[27] E. Zahariev and J. McPhee, *Stabilization of multiple constraints in multibody dynamics using optimization and a pseudo-inverse matrix*, Math. Comput. Model. Dyn. Syst. 9 (4) (2003), pp. 417–435. doi:10.1076/mcmd.9.4.417.27898.

[28] O. Brüls, E. Lemaire, P. Duysinx, and P. Eberhard, *Optimization of Multibody Systems and Their Structural Components, in Multibody Dynamics*, Springer, Dordrecht, (2011), pp. 49–68.

[29] V. Gufler, E. Wehrle, and R. Vidoni, *Multiphysical Design Optimization of Multibody Systems: Application to a Tyrolean Weir Cleaning Mechanism*, The International Conference of IFToMM ITALY, Springer, Cham, (2020), pp. 459-467.

[30] M. Shim and J.-H. Kim, *Design and optimization of a robotic gripper for the FEM assembly process of vehicles*, Mech. Mach. Theory. 129 (2018), pp. 1–16. doi:10.1016/j.mechmachtheory.2018.07.006.

[31] D. Pan, F. Gao, and Y. Miao, *Dynamic research and analyses of a novel exoskeleton walking with humanoid gaits*, J. Mech. Eng. Sci. 228 (2014), pp. 1501–1511.

[32] Y. Miao, F. Gao, and D. Pan, *Prototype design and size optimization of a hybrid lower extremity exoskeleton with a scissor mechanism for load-carrying augmentation*, J. Mech. Eng. Sci. 229 (2015), pp. 155–167.

[33] L. Zhou, Y. Li, and S. Bai, *A human-centered design optimization approach for robotic exoskeletons through biomechanical simulation*, Rob. Auton. Syst. 91 (2017), pp. 337–347. doi:10.1016/j.robot.2016.12.012.

[34] P. Beyl, M. Van Damme, R. Van Ham, B. Vanderborght, and D. Lefeber, *Design and control of a lower limb exoskeleton for robot-assisted gait training*, Appl. Bionics Biomech. 6 (2) (2009), pp. 229–243. doi:10.1155/2009/580734.

[35] S.S. Kawale and M. Sreekumar, *Design of a wearable lower body exoskeleton mechanism for shipbuilding industry*, Procedia. Comput. Sci. 133 (2018), pp. 1021–1028. doi:10.1016/j.procs.2018.07.073.

[36] S. Wang, C. Meijneke, and H. Kooij, *Modeling, Design, and Optimization of Mindwalker Series Elastic Joint*, 2013 IEEE 13th International Conference on Rehabilitation Robotics (ICORR), pp. 1-8, (2013), pp. 1-8, IEEE, doi:10.1109/ICORR.2013.6650381.

[37] S.A.A Moosavian, M.R. Mohamadi, and F. Absalan, *Augmented Modeling of a Lower Limb Assistant Robot and Human Body*, 2018 6th RSI International Conference on Robotics and Mechatronics (IcRoM), (2018), pp. 337-342, Tehran, Iran, IEEE.

[38] M. Nabipour and S.A.A. Moosavian, *Dynamics modeling and performance analysis of RoboWalk*, 2018 6th RSI international conference on robotics and mechatronics (IcRoM), Tehran, Iran, (2018) IEEE.

[39] A.A.M. Nasrabadi, F. Absalan, S. Ali, and A. Moosavian, *Design, kinematics and dynamics modeling of a lower-limb walking assistant robot*, 2016 4th International Conference on Robotics and Mechatronics (ICROM), Tehran, Iran, (2016) IEEE.

[40] M.R. Mohamadi, V. Akbari, O. Mahdizadeh, M. Nabipour, S. Ali, and A. Moosavian, *Simulation analysis of human-RoboWalk augmented model*, Proceedings of the 7th RSI International Conference on Robotics and Mechatronics (IcRoM 2019), 20–21 November 2019, Tehran, Iran, IEEE, .

[41] S.A.A. Moosavian and M. Nabipour, *RoboWalk: Comprehensive augmented dynamics modeling and performance analysis*, Int. J. Rob. Theory. Appl. 5 (1) (2019), pp. 63-72.

[42] R. Featherstone, *Rigid Body Dynamics Algorithms*, 1 ed., Springer US, 2008.

[43] J.J. Craig, *Introduction to Robotics: Mechanics and Control*, Pearson/Prentice Hall, 2005.

[44] B. Siciliano and O. Khatib, *Springer Handbook of Robotics*, Springer International Publishing, 2016.

[45] J. Ashihara, Y. Ikeuchi, T. Fujihira, Y. Nishimura, H. Kudoh, Y. Hiki, M. Shishido, T. Noda, H. Matsuda, Y. Matsuoka, R. Yokoyama, E. Ninomiya, and T. Koshiishi, *Walking Assist Device*, U. Patent, ed., Honda Motor Co., Ltd, Tokyo Japan, 2012.

[46] W. van Dijk, T. Van de Wijdeven, M.M. Holscher, R. Barents, R. Könemann, F. Krause, and C.L. Koerhuis, *Exobuddy - A Non-Anthropomorphic Quasi-Passive Exoskeleton for Load Carrying Assistance, 2018 7th IEEE International Conference on Biomedical Robotics and Biomechatronics (Biorob)* , (2018), pp. 336-341, IEEE.

[47] F. Parietti, K.C. Chan, B. Hunter, and H.H. Asada, *Design and Control of Supernumerary Robotic Limbs for Balance Augmentation, 2015 IEEE International Conference on Robotics and Automation (ICRA),* (2015), pp. 5010-5017, IEEE.

## Appendix

*Notation*

Spatial acceleration
RoboWalk knee joint forces and torque
Forces in RoboWalk ankle
$d_{f\text{-}CoG}$Distance between ankle joint and foot's CoG
$d_{s\text{-}CoG}$Distance between knee joint and shank's CoG
External force exerted on body i
$F$External (foot contact) force
Forces applied to user by RoboWalk
$F_{contact}$Force applied to the foot
$F_x$Horizontal force exerted to the human user
$G$Matrix of gravity effect
$J$Jacobian matrix
LThe magnitude of line drawn between RoboWalk knee and user CoG
$LL_1$, $LL_2$The magnitude of line drawn between RoboWalk knee and first and second bearing
$M(q)$Mass matrix
upper and lower link of RoboWalk
$M_{contact}$Moment applied to the foot
$p$Portion of the bodyweight that the robot supports
rotational motion of pelvis with respect to the inertia.
rotational motion of trunk, hip joints and knee joints with respect to the parent limb
RRoboWalk seat arc radius
Left and right RoboWalk motor torques
Spatial velocity
$V$Matrix of Coriolis and centrifugal and gyration effect
Approximate CoG of user
Coordinate of robot ankle

pelvis reference frame movements with respect to inertia frame
*Greek symbols*
RoboWalk assistive force angles with respect to horizon
Human joint torque
λParent
First robot’s link angle with respect to vertical
Second robot’s link angle with respect to vertical
Angles between L- LL1 and L-LL2
Angle between L and R
The angle between both bearings
*Subscript*
$_{A}$Ankle
$_{H}$Hip
$_{K}$Knee
$_{t}$Trunk
$_{p}$Pelvis
$_{L}$Left
$_{R}$Right
*Abreviations*
CoGCenter of Gravity
DoFDegrees of Freedom
DSPDouble Support Phase
EoMEquations of Motion
FBDFree Body Diagram
FRFFloor Reaction Force
HMILHuman Model In the Loop
HRIHuman-Robot Interaction
PSOParticle Swarm Optimization
RNEARecursive Newton Euler Algorithm
SSPSingle Support Phase